\documentclass[journal,12pt,draftclsnofoot,onecolumn]{IEEEtran}
\IEEEoverridecommandlockouts
\ifCLASSINFOpdf
\else
\fi
\usepackage{graphicx}
\usepackage{bm}
\usepackage{setspace}
\usepackage{diagbox}
\usepackage{flushend}
\usepackage{amssymb}
\usepackage{enumerate}
\usepackage{amsthm}
\usepackage{amsmath}
\usepackage{amsfonts}
\usepackage{subfigure}
\usepackage{cite}
\usepackage{algorithm,algpseudocode,paralist}
\usepackage{colortbl,booktabs}
\usepackage{algorithm}
\usepackage{algpseudocode}

\theoremstyle{definition}

\theoremstyle{remark}

\makeatletter
\def\therule{\makebox[\algorithmicindent][l]{\hspace*{.5em}\vrule height .75\baselineskip depth .25\baselineskip}}%

\newtoks\therules
\therules={}
\def\appendto#1#2{\expandafter#1\expandafter{\the#1#2}}
\def\gobblefirst#1{
	#1\expandafter\expandafter\expandafter{\expandafter\@gobble\the#1}}%
\def\LState{\State\unskip\the\therules}
\def\pushindent{\appendto\therules\therule}%
\def\popindent{\gobblefirst\therules}%
\def\printindent{\unskip\the\therules}%
\def\printandpush{\printindent\pushindent}%
\def\popandprint{\popindent\printindent}%

\algdef{SE}[WHILE]{While}{EndWhile}[1]
{\printandpush\algorithmicwhile\ #1\ \algorithmicdo}
{\popandprint\algorithmicend\ \algorithmicwhile}%
\algdef{SE}[FOR]{For}{EndFor}[1]
{\printandpush\algorithmicfor\ #1\ \algorithmicdo}
{\popandprint\algorithmicend\ \algorithmicfor}%
\algdef{S}[FOR]{ForAll}[1]
{\printindent\algorithmicforall\ #1\ \algorithmicdo}%
\algdef{SE}[LOOP]{Loop}{EndLoop}
{\printandpush\algorithmicloop}
{\popandprint\algorithmicend\ \algorithmicloop}%
\algdef{SE}[REPEAT]{Repeat}{Until}
{\printandpush\algorithmicrepeat}[1]
{\popandprint\algorithmicuntil\ #1}%
\algdef{SE}[IF]{If}{EndIf}[1]
{\printandpush\algorithmicif\ #1\ \algorithmicthen}
{\popandprint\algorithmicend\ \algorithmicif}%
\algdef{C}[IF]{IF}{ElsIf}[1]
{\popandprint\pushindent\algorithmicelse\ \algorithmicif\ #1\ \algorithmicthen}%
\algdef{Ce}[ELSE]{IF}{Else}{EndIf}
{\popandprint\pushindent\algorithmicelse}%
\algdef{SE}[PROCEDURE]{Procedure}{EndProcedure}[2]
{\printandpush\algorithmicprocedure\ \textproc{#1}\ifthenelse{\equal{#2}{}}{}{(#2)}}%
{\popandprint\algorithmicend\ \algorithmicprocedure}%
\algdef{SE}[FUNCTION]{Function}{EndFunction}[2]
{\printandpush\algorithmicfunction\ \textproc{#1}\ifthenelse{\equal{#2}{}}{}{(#2)}}%
{\popandprint\algorithmicend\ \algorithmicfunction}%
\makeatother

\begin{document}
\begin{titlepage}
\begin{center}
\vspace*{-2\baselineskip}
\begin{minipage}[l]{7cm}
\flushleft
\includegraphics[width=2 in]{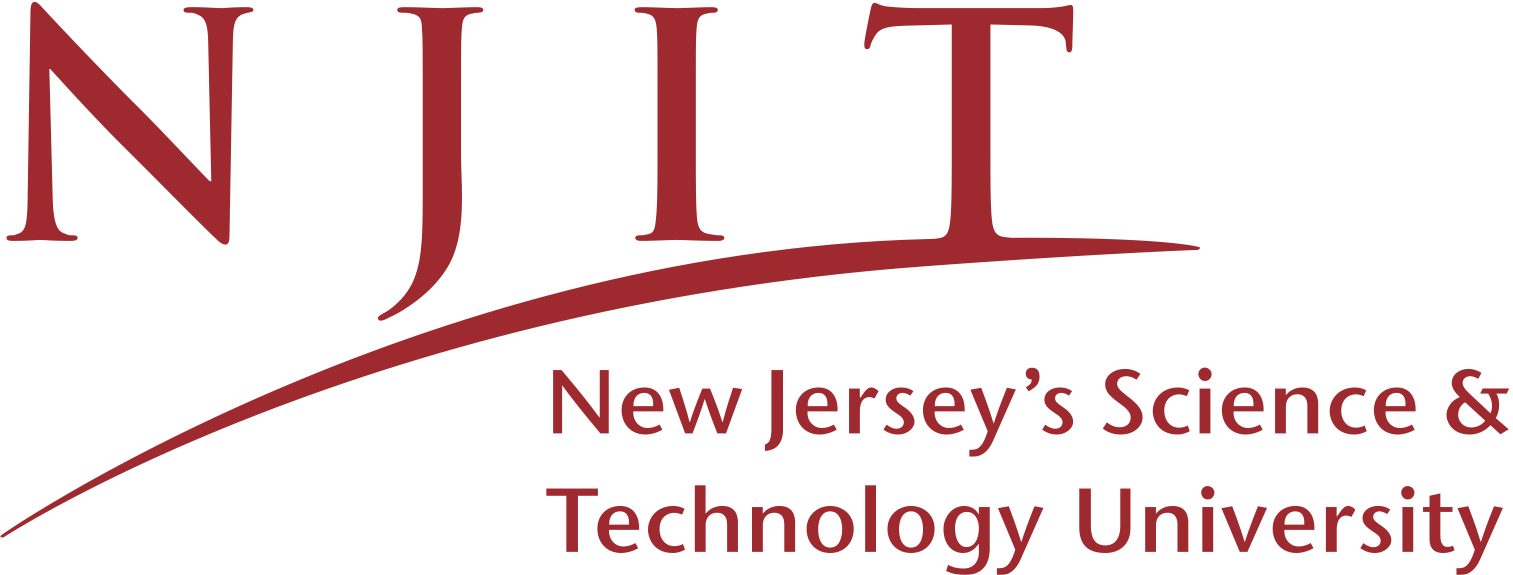}
\end{minipage}
\hfill
\begin{minipage}[r]{7cm}
\flushright
\includegraphics[width=1 in]{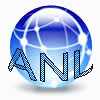}%
\end{minipage}

\vfill

\textsc{\LARGE Indoor Localization by Fusing a Group of Fingerprints Based on Random Forests
}\\

\vfill
\textsc{\LARGE XIANSHENG GUO \\[4pt]
\LARGE NIRWAN ANSARI\\[4pt]
\LARGE HUIYONG LI\\[4pt]}
\vfill
\textsc{\LARGE TR-ANL-2016-004\\[12pt]
\LARGE December 22, 2016}\\[1.5cm]
\vfill
{ADVANCED NETWORKING LABORATORY\\
 DEPARTMENT OF ELECTRICAL AND COMPUTER ENGINEERING\\
 NEW JERSY INSTITUTE OF TECHNOLOGY}
\end{center}
\end{titlepage}

\section{abstract}
Indoor localization based on SIngle Of Fingerprint (SIOF) is rather susceptible to the changing environment, multipath, and non-line-of-sight (NLOS) propagation. Building SIOF is also a very time-consuming process. Recently, we first proposed a GrOup Of Fingerprints (GOOF) to improve the localization accuracy and reduce the burden of building fingerprints. However, the main drawback is the timeliness. In this paper, we propose a novel localization framework by Fusing A Group Of fingerprinTs (FAGOT) based on random forests. In the offline phase, we first build a GOOF from different transformations of the received signals of multiple antennas. Then, we design multiple GOOF strong classifiers based on Random Forests (GOOF-RF) by training each fingerprint in the GOOF. In the online phase, we input the corresponding transformations of the real measurements into these strong classifiers to obtain multiple independent decisions. Finally, we propose a Sliding Window aIded Mode-based (SWIM) fusion algorithm to balance the localization accuracy and time. Our proposed approaches can work better in an unknown indoor scenario. The burden of building fingerprints can also be reduced drastically. We demonstrate the performance of our algorithms through simulations and real experimental data using two Universal Software Radio Peripheral (USRP) platforms.
\section{IEEEkeywords}
GrOup Of Fingerprints (GOOF), Sliding Window aIded Mode-based (SWIM) fusion, multiple antennas, USRP, Random Forests.
\section{Introduction}
%
%
%
%
\label{sec:intro}
\IEEEPARstart{E}{mergence} of location-based service and applications has led to a growing demand for space sensing and localization \cite{guo2016accurate, huang2015green}. Although Global Positioning System (GPS) has gained great success in many outdoor localization fields, such as commercial, personal, and military applications, it does not perform effectively in complex indoor environments owing to the disability of GPS signals to penetrate in-building materials. Therefore, precise indoor localization is well sought and critical for a wide range of applications.

Indoor localization environment consists of severe multipath and non-line-of-sight (NLOS) between the  transmitter and receiver. In addition, the changing environment resulted from moving people and closing/opening of doors and windows, presents a big challenge for indoor localization. These factors degenerate the performance of some range-based indoor localization approaches \cite{wang2014nlos, Wang2013A}.
The fingerprint-based approach does not need to estimate the distance between the transmitter and receiver. It achieves better performance than the range-based approach in a complex indoor environment. However, most of the existing fingerprint-based approaches are based on a SIngle Of Fingerprint (SIOF), such as received signal strength (RSS). The major challenge of RSS is its fluctuation with time and changing environment. So, RSS shows low accuracy and poor robustness in practice. Other SIOFs, including channel impulse response (CIR) \cite{nerguizian2006geolocation, jin2010indoor},  signal strength difference (SSD) \cite{mahtab2013ssd, guo2014robust}, signal subspace \cite{tsuji2006indoor, ikeda2007effects}, power delay doppler profile (PDDP) \cite{oktem2010power}, can improve the accuracy of indoor localization to some extent. All in all, they all belong to the SIOF-based localization framework, which cannot work well in an unknown indoor environment because it only uses little information about indoor environment.

Another drawback of the SIOF-based localization approach is the big burden of building fingerprint.  To reduce this burden, some fingerprint building strategies have been proposed, including crowdsourcing \cite{wu2015smartphones}, matrix completion (MC) \cite{nikitaki2012efficient}, compressive sensing (CS) \cite{gu2015reducing}, and others \cite{talvitie2015distance}.  These techniques can alleviate the fingerprint building burden from different viewpoints. However, the  performance of rebuilding fingerprints may decrease as the number of samples in the original fingerprint decreases.

Recently, we first creatively proposed a GrOup Of Fingerprints (GOOF) based localization framework \cite{guo2016localization}. GOOF can overcome the drawbacks resulted from SIOF. Based on the constructed GOOF, after training multiple GOOF-AadaBoost classifiers, we used MUltiple Classifiers mUltiple Samples (MUCUS) to  determine the final location prediction. MUCUS can yield higher accurate results. However, the main drawback of MUCUS is the timeliness.

\begin{figure}[htbp]
  	\centering
  	\includegraphics[width=0.65\columnwidth]{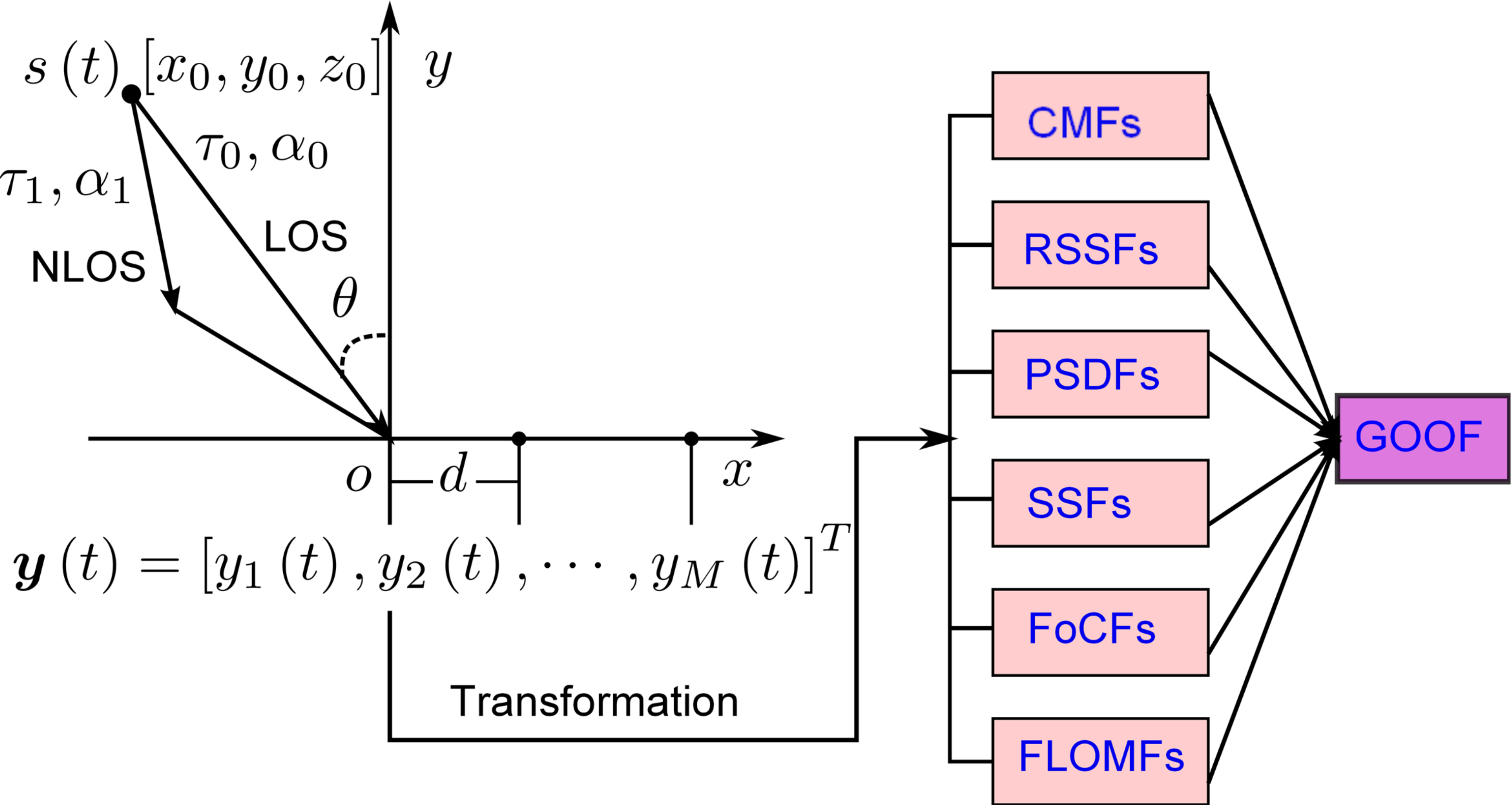}
  	\caption{The proposed GOOF building sketch by using multiple antennas.}
  	\label{fig:fusion_localization_framework}
  \end{figure}

In this study, we propose a novel localization framework by Fusing A Group Of fingerprinTs (FAGOT) based on random forests. The new proposed GOOF is composed of six different kinds of fingerprints, namely, RSS fingerprints (RSSFs), power spectral density fingerprints (PSDFs), covariance matrix fingerprints (CMFs), and signal subspace fingerprints (SSFs), fourth-order cumulant fingerprints (FoCFs), and fractional low order moment fingerprints (FLOMFs),  which can be
obtained by different transformations of the received signals ${\bm{y}}\left(t\right)$
of multiple antennas, as illustrated in Fig. \ref{fig:fusion_localization_framework}. Each fingerprint in the GOOF has its special function. Among them, RSSFs reflect the distance between a transmitter and a receiver; SSFs are robust to multipath propagation \cite{tsuji2006indoor}; CMFs, FoCFs, and FLOMFs are robust to Gaussian, color, and impulse noise, respectively; PSDFs describe the distribution of signal power in the frequency domain and have been used as an efficient fingerprint in many fields \cite{suski2008using}. In a real indoor localization scenario, the types of noise and environment are changing and cannot be predicted
in advance, and we cannot know which fingerprint can work better in an unknown indoor scenario.
Based on the constructed GOOF, we design a GOOF multiple classifiers based on Random Forests (GOOF-RF) to train fingerprints in the GOOF. Finally,
we localize the target by inputting the corresponding transformations of the online data into the strong classifiers.
A fast fusion strategy, referred to as the Sliding Window Aided Mode-based (SWIM) fusion algorithm, is proposed to refine a better location estimation.

 Our proposed localization framework consists of two phases: an offline phase, which includes GOOF building and GOOF-RF training, and an online localization phase, which includes GOOF-RF testing and SWIM fusion, as summarized below.
 \begin{itemize}
  	\item  \textbf{The offline phase} \\
  	\begin{enumerate}  [a).]
  		\item \textbf{GOOF building:} Assume that we have $\mathcal{Q}$ grids in an unknown indoor environment, the received array with $M$ antennas is deployed at the origin, and $L$ snapshots ${\bm{Y}}=\left[{\bm{y}}\left(1\right),{\bm{y}}\left(2\right),\cdots,{\bm{y}}\left(L\right)\right]$ of size $M \times L$ are collected at each grid. Then, we can build the GOOF by using ${\bm{Y}}$ with different transformations. The $\mathcal{Q}$ different labels are also added into the constructed GOOF for further classification.
  		\item \textbf{GOOF-RF training:}
  		After obtaining the GOOF, we divide  each fingerprint in the GOOF into two parts, one to train the GOOF multiple classifiers based on random forests (GOOF-RF) and the other to test these classifiers. Assume that we have $\mathcal{H}$ different kinds of fingerprints in our constructed GOOF; then, we can train $\mathcal{H}$ strong classifiers by using random forests.
  	\end{enumerate}
  	\item \textbf{The online phase} \\
  	\begin{enumerate}[a).]
  		\item \textbf{GOOF-RF testing:}
  		Assume that we can obtain $\mathcal{Z} \left(\mathcal{Z}\ll L\right)$ samples of each fingerprint for testing. First, we input all the testing data to the trees in the $\mathcal{H}$ random forest strong classifiers. Then, each classifier outputs an $\mathcal{Z} \times {1}$ prediction vector ${\bm{b}}_{\gamma},\left(\gamma=1,\cdots,\mathcal{H}\right)$. The total prediction matrix is ${\bm{B}}=\left[{\bm{b}}_1,\cdots,{\bm{b}}_{\mathcal{H}}\right]$, as depicted in Fig. \ref{fig:MCMSF}.
  		\item \textbf{SWIM fusion:}
  		Based on $\bm{B}$, we choose a rectangle sliding window of length $\mathcal{W}$ to provide fast prediction.  Our proposed SWIM fusion algorithm can optimize the balance between  the localization accuracy and speed, which is very attractive in the real environment.
  	\end{enumerate}
 \end{itemize}

 The proposed localization algorithm can synthesize not only  different predictions of ${\mathcal{H}}$ strong classifiers but also the  predictions of each strong classifier with different samples. The adopted sliding window strategy readily balances the localization speed and accuracy  simultaneously. The main contributions of this work are summarized below:
  \begin{itemize}
  	\item As compared with the SIOF-based framework, our proposed FAGOT localization framework can achieve better performance in an unknown indoor environment in that it can fuse the predictions of different classifiers. Theoretically, the more different kinds of fingerprints in the GOOF, the higher localization accuracy of FAGOT.
  	\item The proposed GOOF-RF training and testing algorithms are simple to implement, and less susceptible to overfitting without tuning a bunch of parameters. The testing time is shorter and easier to realize.
  	\item The proposed SWIM fusion
  	algorithm can balance the localization speed and  accuracy simultaneously by combining the predictions of multiple classifiers and different samples with a sliding window. The localization time could be shortened to $\frac{1}{\mathcal{Z}-\mathcal{W}+1}$ of MUCUS with $\mathcal{W}$ being the length of a sliding window.
  	\item The fingerprints building burden is reduced efficiently because the GOOF building strategy can obtain different kinds of fingerprints based on the measurements of received antennas. Specifically, in case of building the same number of fingerprints, the building burden of the GOOF is ${1 \mathord{\left/
  			{\vphantom {1 H}} \right.
  			\kern-\nulldelimiterspace} \mathcal{H}}$ of that of the SIOF-based approaches. Meanwhile, the localization accuracy can be improved remarkably.
  \end{itemize}

  \section{Related Work}
\label{sec:relate}
The existing indoor localization techniques using wireless sensor networks (WSN), wireless local area networks(WLAN), radio frequency identification (RFID) technology, light visible communication, and others \cite{kim2013indoor, zhang2010wireless, guo2009low}, have limited ability in coping with complex multipath, NLOS, and changing environment in indoor scenarios. In the past few decades, array signal processing has gained tremendous achievements in outdoor target identification, direction finding, and beamforming \cite{guo2009low, guo2010source}.
As the development of antenna technique and high speed baseband processing integrated circuit (IC), small array processing platforms, such as USRP, have been used in many fields  based on software defined radio (SDR)  technology \cite{friedman2009angle}. Hence,  indoor localization using small platform with multiple antennas becomes feasible and has been a hot research subject.  Kleisouris {\textit{et al.}} \cite{kleisouris2010empirical} provided an experimental evaluation of the localization performance under multiple antennas and showed that the localization accuracy can be improved greatly by employing multiple low-cost antennas regardless of whether fingerprint matching, statistical maximum likelihood estimation, or multilateration is used. Note that this conclusion was drawn by only using RSS of multiple antennas as the metric.
	
Recently, random forest, as one of the most popular machine learning techniques, has been studied widely in many fields \cite{sonka2014image}.  The random forest is unexcelled in accuracy and runs efficiently on large databases. It can handle thousands of input variables without variable deletion as well as give estimates on which variables are important in the classification. Meanwhile, it can effectively estimate missing data and maintain accuracy when a large proportion of the data are missing. Although random forest has achieved tremendous success in regression and classification problems, it is seldom studied in indoor localization. Calderoni {\textit{et al.}} \cite{calderoni2015indoor} studied an indoor localization approach by using random forest classifiers based on RFID technology. Jedari {\textit{et al.}} \cite{jedari2015wi} proposed a random forest based localization approach in the WLAN environment. All these approaches belong to the SIOF-based indoor localization framework. To make full use of multiple fingerprints, we first creatively proposed a GOOF based localization framework in \cite{guo2016localization}. The constructed GOOF is trained as multiple strong classifiers by using AdaBoost. A multiple classifiers multiple samples (MUCUS) fusion algorithm was proposed to fuse the predictions of these strong classifiers with multiple samples. The robustness and accuracy of \cite{guo2016localization}  are good enough but the speed of localization is slightly slow. In this work, we propose a novel FAGOT indoor localization framework based on random forests to overcome the above drawbacks as follows.

\section{METHODOLOGY}
\label{sec:model}
\subsection{Signal Model}
\label{sec:subsec:signal_model}
Consider an indoor environment deployed with a uniform linear array (ULA) in which $M$ antenna elements are equally spaced apart, with an inter-distance of $d$, as shown in Fig. \ref{fig:fusion_localization_framework}. Let $y_m\left(t\right)$ denote the received signal at the $m$th antenna element with channel gain ${\alpha}_i$, delay ${\tau}_i$, and angle-of-arrival (AoA) ${\theta}_i$. Note that the received signal of each path consists
of an enormous number of unresolvable signals received
around the mean of AoA in each element in a complex
indoor scenario. A signal $s\left(t\right)$ is transmitted from the location of $\bm{x}=\left[x_0,y_0,z_0\right]^T$. A vector of the received signals ${\bm{y}}\left(t\right)=\left[y_1\left(t\right),y_2\left(t\right),\cdots,y_M\left(t\right)\right]^T$ in the ULA can be expressed as \cite{cho2010mimo, guo2009low}
\begin{eqnarray}
\label{eq:signal_model1}
{\bm{y}}\left(t\right)={\sum\limits_{i=0}^{I-1}}\alpha_i {\bm{a}}\left(\theta_i\right)s\left(t-\tau_i\right)+{\bm{n}}\left(t\right),
\end{eqnarray}
where $I$ denotes the number of paths received by each antenna element and ${\bm{a}}$ is an array steering vector. The location $\bm{x}$ of  the transmitted signal $s\left(t\right)$ is to be estimated. The unknown noise vector ${\bm{n}}\left(t\right)=\left[n_1\left(t\right),n_2\left(t\right),\cdots,n_M\left(t\right)\right]^T$  with $n_m\left(t\right)$ being the noise of the $m$th antenna element. The array steering vector is defined as
${\bm{a}}\left(\theta\right)=\left[a_1\left(\theta\right),a_2\left(\theta\right),\cdots,a_M\left(\theta\right)\right]^T$, where its $m$th element is
\begin{eqnarray}
\label{eq:signal_model2}
a_m\left(\theta\right)=f_m\left(\theta\right)e^{-j2\pi\left(m-1\right)\left(d/\lambda\right)\sin\theta},
\end{eqnarray}
where $f_m\left(\theta\right)$ denotes a complex field pattern of the $m$th array element and $\lambda$ is the carrier wavelength. The received signal in Eq. (\ref{eq:signal_model1}) can be expressed in the following integral form:
\begin{eqnarray}
\label{eq:signal_model3}
{\bm{y}}\left(t\right)=
\iint {\bm{a}}\left(\theta\right)h\left(\theta,\tau\right)s\left(t-\tau\right)d{\tau}d\theta+{\bm{n}}\left(t\right),
\end{eqnarray}
where $h\left(\theta,\tau\right)$ represents the channel as a function of the azimuth-delay spread (ADS). The average power azimuth-delay spectrum (PADS) is given by
\begin{eqnarray}
\label{eq:signal_model4}
P\left(\theta,\tau\right)=E\left\{{\sum\limits_{i=1}^{I}}\left|\alpha_i\right|^2\delta\left(\theta-\theta_i,\tau-\tau_i\right)\right\},
\end{eqnarray}
where $E\left\{\cdot\right\}$ is the expectation operator and $\delta\left(\cdot\right)$ is the Dirac deta function.
The  central angular of arrival (CAoA) $\theta_0$  and angular spread (AS) $\sigma_A$ are defined as
\begin{eqnarray}
\label{eq:signal_model5}
\left\{ \begin{array}{l}
\theta_0=\int\theta P_A\left(\theta\right) d\theta,\\
\sigma_A=\sqrt{\int\left(\theta-\theta_0\right)^2 P_A\left(\theta\right)d\theta},
\end{array} \right.
\end{eqnarray}
where $P_A\left(\theta\right)=\int P\left(\theta,\tau\right)d\tau$ is the power angular spectrum (PAS).

Similarly, the average delay spread (ADS) and delay spread (DS) can be given by
\begin{eqnarray}
\label{eq:signal_model6}
\left\{ \begin{array}{l}
\tau_0=\int\theta P_D\left(\tau\right) d\tau,\\
\sigma_D=\sqrt{\int\left(\tau-\tau_0\right)^2 P_D\left(\tau\right)d\tau},
\end{array} \right.
\end{eqnarray}
where $P_D\left(\tau\right)=\int P\left(\theta,\tau\right)d\theta$ is the power delay spectrum (PDS).
The indoor localization problem using ULA is to estimate $\bm{x}$ from the $L$ measurements of ${\bm{y}}\left(t\right)$.
\subsection{GOOF Building}
\label{sec:subsec:FsG_building}
Here, we address how to build our proposed GOOF from the received signals ${\bm{y}}\left(t\right)$ by using $L$ snapshots. Assume that we divide the indoor environment into $\mathcal{Q}$ grids with equal spacing. The signal $s\left(t\right)$ is transmitted from one antenna located at the $q$th grid, and the received signals vector of $M$ antenna elements at time $t$ is denoted by ${\bm{y}}^q\left(t\right)$.
\begin{itemize}
	\item \textbf{Covariance matrix fingerprints (CMFs)} \\
	We can estimate the covariance matrix by using $L$ snapshots at the $q$th grid without any knowledge of noise distributions as follows:
	\begin{eqnarray}
	\label{eq:FsG_building1}
	{\bm{\hat{R}}}^q=\frac{1}{L}\sum\limits_{t=1}^L {\bm{y}}^q\left(t\right){{\bm{y}}^{q}\left(t\right)}^H.
	\end{eqnarray}
	Note that  the estimated covariance matrix (\ref{eq:FsG_building1}) can be expressed as
	\begin{eqnarray}
	\label{eq:FsG_building2}
	{\bm{\hat{R}}}^q = \left[ {\begin{array}{*{20}{c}}
		{r\left( 0 \right)}&{r\left( { - 1} \right)}& \cdots &{r\left( { - M+1} \right)}\\
		{{r}\left( {1} \right)}&{r\left( 0 \right)}& \cdots &{r\left( { - M + 2} \right)}\\
		\vdots & \vdots & \ddots & \vdots \\
		{{r}\left( {  M-1} \right)}&{{r}\left( { M -2} \right)}& \cdots &{r\left( 0 \right)}
		\end{array}} \right].
	\end{eqnarray}
	The $\left(i,j\right)$th entry of (\ref{eq:FsG_building2}) is the correlation between the outputs of the $i$th and  $j$th antennas. We can estimate the RSS from (\ref{eq:FsG_building2}) as follows.
	\item \textbf{RSS fingerprints (RSSFs)} \\
	It is well known that the $i$th diagonal element of the estimated covariance matrix $r\left(0\right)$ in Eq. (\ref{eq:FsG_building2}) denotes the autocorrelation of the received signals $y_i\left(t\right)$ of the $i$th antenna element, i.e.,
	\begin{eqnarray}
	\label{eq:FsG_building3}
	r_i\left(0\right)=\frac{1}{L}\sum\limits_{t=1}^L {{y}}_i\left(t\right){{{y}}_i\left(t\right)}=\frac{1}{L}\sum\limits_{t=1}^L\left|y_i\left(t\right)\right|^2.
	\end{eqnarray}
	So, we can build the RSS fingerprints by taking the diagonal elements of (\ref{eq:FsG_building2}), i.e.,
	\begin{eqnarray}
	\label{eq:FsG_building4}
	\textbf{RSS}^q=\left[r_1^q\left(0\right),r_2^q\left(0\right),\cdots,r_M^q\left(0\right)\right]^T={\rm{diag}}\{{\bm{\hat{R}}}^q\},
	\end{eqnarray}
	where $\rm{diag\{\cdot\}}$ is the operator of extracting the diagonal elements of a matrix. In comparing with Eqs. (\ref{eq:FsG_building2}) and (\ref{eq:FsG_building4}), it is remarkable that the CMFs can offer more information about the indoor channel than that of the RSSFs because the CMFs have much correlation information among antenna elements. So, we have enough reasons to believe that the CMFs yield a more accurate location estimate than that of the RSSFs.
	\item \textbf{Power spectral density fingerprints (PSDFs)}\\
	The normalized PSD can be calculated by
	\begin{eqnarray}
	\label{eq:psd}
	\textbf{PSD}^q\left(m,k\right)=\frac{\left|Y_m\left(k\right)\right|^2}{\sum\nolimits_{k=1}^K \left|Y_m\left(k\right)\right|^2 },
	\end{eqnarray}
	where $Y_m\left(k\right)$ is a sequence of complex Discrete Fourier Transform (DFT) coefficients for the  received signal sequence ${y}_m\left(t\right)$ of the $m$th antenna, which is given by
	\begin{eqnarray}
	\label{eq:psd1}
	Y_m\left( k \right) = \frac{1}{{{L}}}\sum\limits_{t = 1}^{{L}} {y_m\left( t \right)\exp \left[ {\frac{{ - 2\pi j}}{{{L}}}\left( {t - 1} \right)\left( {k - 1} \right)} \right]}
	\end{eqnarray}
	in which $L$ is the DFT length and $K$ is the point number in the frequency domain.
	\item \textbf{Signal subspace fingerprints (SSFs)} \\
	By taking eigen-decomposition (ED) of the estimated covariance matrix, we have
	\begin{eqnarray}
	\label{eq:FsG_building5}
	{{\bm{R}}^q} = \left[ {\begin{array}{*{20}{c}}
		{{{\bm{U}}_s^ q}}&{{{\bm{U}}_n^q}}
		\end{array}} \right]\left[ {\begin{array}{*{20}{c}}
		{{{\bm{\Sigma}} _s^q}}&{}\\
		{}&{{{\bm{\Sigma}} _n^q}}
		\end{array}} \right]\left[ {\begin{array}{*{20}{c}}
		{{{\bm{U}}_s^q}^H}\\
		{{{\bm{U}}_n^q}^H}
		\end{array}} \right],
	\end{eqnarray}
	where ${\bm{\Sigma}} _s^q$ is the signal subspace corresponding to the $k$ largest eigenvalues whose elements are  the diagonal elements of the diagonal matrix ${{\bm{\Sigma}} _s^q}$; ${\bm{U}}_n^q$ is the noise subspace, which corresponds to the $M-k$ small eigenvalues. Signal subspace methods are empirical linear methods for dimensionality reduction and noise reduction. They have also been demonstrated to be robust to multipath propagation in indoor localization \cite{tsuji2006indoor}. Note that we just build the signal subspace fingerprints by taking the first column of ${\bm{U}}_s^q$ instead of finding the $k$ columns of ${\bm{U}}_s^q$ for simplicity.
	\item \textbf{Fourth-order cumulant fingerprints (FoCFs)} \\
	The FoC of the received signals ${\bm{y}}\left(t\right)$ can be given by
	\begin{eqnarray}
	\label{eq:FsG_building6}
	\begin{array}{l}
	{{\bm{C}}_{4,y}^q}={\rm{ cum}}\left\{ {{y_{{k_1}}},{y_{{k_2}}},y_{{k_3}}^ * ,y_{{k_4}}^ * } \right\}\\
	{\kern 1pt} {\kern 1pt}  = {E}\left\{ {{y_{{k_1}}}{y_{{k_2}}}y_{{k_3}}^ * y_{{k_4}}^ * } \right\} - { E}\left\{ {{y_{{k_1}}}y_{{k_3}}^ * } \right\}{ E}\left\{ {{y_{{k_2}}}y_{{k_4}}^ * } \right\}\\
	{\kern 1pt}   - { E}\left\{ {{y_{{k_1}}}y_{{k_4}}^ * } \right\}{ E}\left\{ {{y_{{k_2}}}y_{{k_3}}^ * } \right\} - { E}\left\{ {{y_{{k_1}}}{y_{{k_2}}}} \right\}{ E}\left\{ {y_{{k_3}}^ * y_{{k_4}}^ * } \right\}
	\end{array}
	\end{eqnarray}
	where
	\begin{eqnarray}
	\label{eq:FsG_building7}
	{E}\left\{ {{y_{{k_i}}}{y_{{k_j}}}y_{{k_m}}^ * y_{{k_n}}^ * } \right\} = \frac{1}{{{L}}}\sum\limits_{t = 1}^{{L}} {{y_{{k_i}}}\left( t \right){y_{{k_j}}}\left( t \right)y_{{k_m}}^ * \left( t \right)y_{{k_n}}^ * } \left( t \right)
	\end{eqnarray}
	and
	\begin{eqnarray}
	\label{eq:FsG_building8}
	{ E}\left\{ {{y_{{k_i}}}y_{{k_j}}^ * } \right\} = \frac{1}{{{L}}}\sum\limits_{t = 1}^{{L}} {{y_{{k_i}}}\left( t \right)} y_{{k_j}}^ * \left( t \right).
	\end{eqnarray}
	It is well known that the FoCFs are generally robust to color noise \cite{nikias1993higher}.
	\item \textbf{Fractional low order moments fingerprints (FLOMFs)} \\
	Impulsive noise distorts the signal and causes the degeneration of localization accuracy of source. Studies in \cite{zhong2013particle} have shown that the symmetric alpha-stable (S$\alpha$S) processes are able to model the impulsive noise better. We can calculate the FLOMFs as follows \cite{liu2001subspace}.
	\begin{eqnarray}
	\label{eq:FsG_building9}
	{\bm{C}}_{f,y}^q=E\{y_i\left(t\right)\left|y_k\left(t\right)\right|^{p-2}y_k^*\left(t\right)\}, 1<p<\alpha \le 2,
	\end{eqnarray}
	where $0 < \alpha \le 2$ is the characteristic exponent of an S$\alpha$S processes. Note that when $p=2$, Eq. (\ref{eq:FsG_building9}) is the special case of Eq. (\ref{eq:FsG_building1}). However, for impulse noise, the FLOM is unbounded. The FLOM is a good statistic used to estimate DOAs of sources in array signal processing field.
\end{itemize}

\begin{table}[!t]
	\caption{Transformations on the GOOF. }
	\label{tab:transform1}
	\begin{tabular}{p{1.5cm}p{2.3cm}p{1.5cm}p{1.5cm}}
		\toprule
		Fingerprint   & Transformations        & Dimension before transformation & Dimension after transformation \\ \midrule
		\textbf{ {CMFs}}   & \rm{reshape}, \rm{abs} & $M\times M$                   & $M^2\times 1$                \\
		\textbf{ {PSDFs}}  & \rm{reshape} & $M\times K$                   & $MK\times 1$                 \\
		\textbf{ {FoCFs}}  & \rm{reshape}, \rm{abs} & $M\times M$                   & $M^2\times 1$                \\
		\textbf{ {FLOMFs}} & \rm{reshape}, \rm{abs} & $M\times M$                   & $M^2\times 1$                \\
		\textbf{ {SSFs}}   & \rm{abs}               & $M\times 1$                   & $M\times 1$                  \\
		\textbf{ {RSSFs}}  & none                   & $M\times 1$                   & $M\times 1$                  \\ \bottomrule
	\end{tabular}
\end{table}

So far, we have addressed how to build the GOOF based on the received signals. Note that the dimensions of the six proposed fingerprints in the GOOF are not the same. Except for the RSSFs, the rest of them are complex values. For the complex fingerprints, we just take absolute values of them and drop the phase information, which is sensitive to the noise level. We adjust the dimensions and data types of the constructed GOOF,  as shown in Table. \ref{tab:transform1}. For comparisons, we summarize the GOOF building procedures in Algorithm \ref{ALG:FsGbuilding}. To obtain as many fingerprints as possible at each grid for further random forests classifiers training, we partition the $L$ snapshots into $\mathcal{M}$ groups with each group having ${L \mathord{\left/
		{\vphantom {L M}} \right.\kern-\nulldelimiterspace} {\mathcal{M}}}$ snapshots. We just use the ${L \mathord{\left/
		{\vphantom {L M}} \right.\kern-\nulldelimiterspace} {\mathcal{M}}}$ snapshots to estimate each fingerprint.

It is worth to note that the proposed GOOF building strategy can reduce the fingerprints building burden as compared with the SIOF-based approaches \cite{nikitaki2012efficient, talvitie2015distance, gu2015reducing}. The GOOF building strategy can obtain multiple types of fingerprints with different transformations from the same measurements, while the SIOF building strategies can only obtain one kind fingerprint from the same measurements. Hence, the efficiency of our  GOOF building strategy is much higher than the SIOF-based building strategies. In other words, GOOF can obtain the same localization precision with less fingerprints building time as compared with the SIOF.  Furthermore, the SIOF building strategies can only reduce the fingerprints building burden but not improve the localization accuracy; our GOOF strategy can not only reduce the fingerprints building burden, but also improve the accuracy of localization, which is  very attractive for real application.

\begin{algorithm}[!t]
	\algnewcommand\algorithmicinput{\textbf{Input:}}
	\algnewcommand\algorithmicoutput{\textbf{Output:}}
	\algnewcommand\INPUT{\item[\algorithmicinput]}
	\algnewcommand\OUTPUT{\item[\algorithmicoutput]}
	\caption{\textbf{GOOF building}}
	\label{ALG:FsGbuilding}
	\begin{algorithmic}[1]
		\INPUT{
			\begin{inparaenum}
				\item{The received signals of  ${\bm{y}}\left(t\right), t=1,2,\cdots, L$.}
				\item{The number of grid $\mathcal{Q}$.}
				\item{The location label $q, \left( q=1,2,\cdots,{\mathcal{Q}}\right)$.}
				\item{The initial empty GOOF  $ \textbf{GOOF}=\emptyset$, $\textbf{ {CMFs}}=\emptyset$, $ \textbf{RSSFs}=\emptyset$, $ \textbf{PSDFs}=\emptyset$, $\textbf{SSFs}=\emptyset$, $\textbf{FoCFs}=\emptyset$, $\textbf{FLOMFs}=\emptyset$.}
				\item {The group number $\mathcal{M}$  at each grid.}
			\end{inparaenum}
		}
		\OUTPUT {\begin{inparaenum}
				{\textbf{GOOF}}.
			\end{inparaenum}}
			\For {$q=\{1,\cdots,{\mathcal{Q}}\}$ }
			\For {$k=\{1,\cdots,\mathcal{M}\}$ }
			\LState Calculate ${\bm{\hat{R}}}^q$ by using Eq. (\ref{eq:FsG_building1})
			\LState Calculate $\textbf{RSS}^q$ by using Eq. (\ref{eq:FsG_building4})
			\LState Calculate $\textbf{PSD}^q$ by using Eq. (\ref{eq:psd})
			\LState Calculate ${\bm{U}}_s^q$ by using Eq. (\ref{eq:FsG_building5})
			\LState Calculate ${{\bm{C}}_{4,y}^q}$ by using Eq. (\ref{eq:FsG_building6})
			\LState Calculate ${{\bm{C}}_{f,y}^q}$ by using Eq. (\ref{eq:FsG_building9})
			\LState Transform the GOOF like Table. \ref{tab:transform1}.
			\LState \textbf{ {CMFs}} =\textbf{ {CMFs}} $\cup {\bm{\hat{R}}}^q \cup q$
			\LState \textbf{ {RSSFs}} =\textbf{ {RSSFs}} $\cup \textbf{RSS}^q \cup q$
			\LState \textbf{ {PSDFs}} =\textbf{ {PSDFs}} $\cup \textbf{PSD}^q \cup q$
			\LState \textbf{ {SSFs}} =\textbf{ {SSFs}} $\cup {\bm{U}}_s^q \cup q$
			\LState \textbf{ {FoCFs}} =\textbf{ {FoCFs}} $\cup {{\bm{C}}_{4,y}^q} \cup q$
			\LState \textbf{ {FLOMFs}} =\textbf{ {FLOMFs}} $\cup {{\bm{C}}_{f,y}^q} \cup q$
			\EndFor
			\EndFor
			\LState  \textbf{GOOF}=$\textbf{GOOF}\cup\textbf{ {CMFs}}\cup\textbf{ {RSSFs}}\cup\textbf{ {PSDFs}}\cup\textbf{ {SSFs}}\cup\textbf{ {FoCFs}}\cup\textbf{ {FLOMFs}}$
			\LState\Return  \textbf{GOOF}
		\end{algorithmic}
	\end{algorithm}
	\subsection{GOOF multiple classifiers training and testing  based on random forests (GOOF-RF)}
	\label{sec:subsec:fgpmcboadaboost}

Random forests (RF) are a combination of tree predictors such that each tree depends on the values of a random vector sampled independently and with the same distribution for all trees in the forest \cite{breiman2001random}. The key aspect of random forest is the fact that its component trees are all randomly different from one another. This leads to decorrelation between the individual tree predictions and, in turn, results in improved generalization and robustness. A tree is a collection of nodes and edges organized in a hierarchical structure. Nodes are divided into split nodes and  leaf nodes. All nodes have exactly one incoming edge.
\begin{figure}[!t]
\centering
\includegraphics[width=0.65\columnwidth]{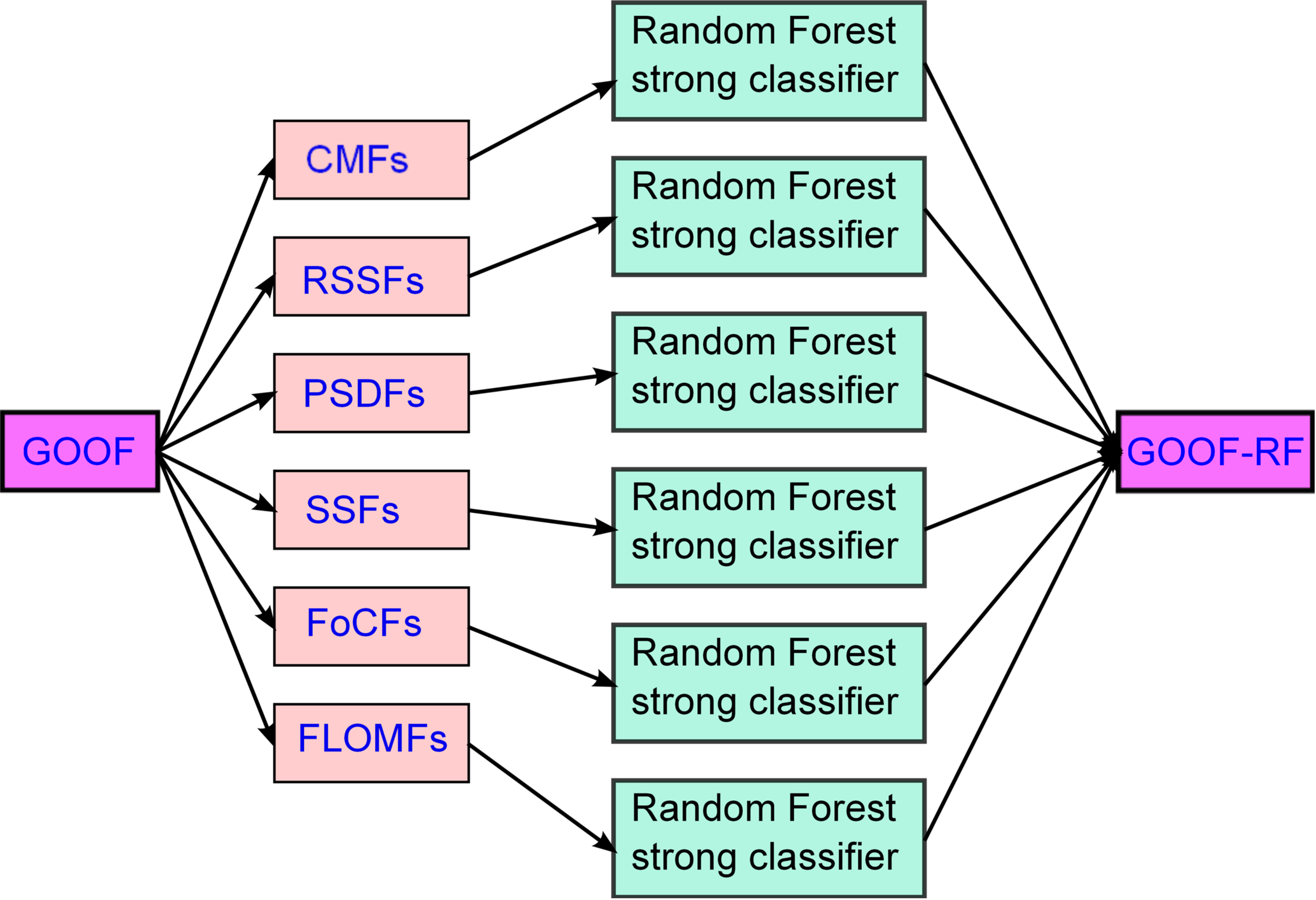}
\caption{The framework of our proposed GOOF-RF training approach.}
\label{fig:PMACF}
\end{figure}	
	 Our proposed GOOF multiple classifiers based on random forests (GOOF-RF) will build multiple strong classifiers from our constructed GOOF. Each strong classifier yields
	 its final location estimation of the target.  We illustrate our proposed GOOF-RF training procedures   in Fig. \ref{fig:PMACF}, which shows that each kind of fingerprints can be trained as a random forest classifier. We summarize the basic principle of random forests as follows.
	
	Let  ${\bm{\zeta}}=\left(\zeta_1,\zeta_2,\cdots,\zeta_{\tilde{d}}\right) \in \mathcal{F}$ be a data vector, where the components $\zeta_i$ represent some attributes of the vector and $\mathcal{F}$ represents the constructed GOOF; ${\tilde{d}}$ is the dimensionality of ${\bm{\zeta}}$.  In our case, $\bm{\zeta}$ represents the different fingerprints vector built in the GOOF. Note that $\tilde{d}$ may vary from different kinds of fingerprints. In general, the feature space $\mathcal{F}$ and $\tilde{d}$ can be very large, especially in the image processing field. Fortunately, we can extract only a small portion of $\tilde{d}$ as $\phi\left(\bm{\zeta}\right)=\left(\zeta_{\phi_1},\zeta_{\phi_2},\cdots,\zeta_{\phi_{m}}\right)\in{\mathcal{F}^{m}}\subset{\mathcal{F}}$, where $m$ denotes the dimensionality of the subspace and $\phi_i\in[1,{\tilde{d}}]$ denotes the selected dimensions. In general, $m<<\tilde{d}$. Each node has a test function with binary outputs
	\begin{eqnarray}
	\label{eq:goof-rf-eq1}
	h'\left({\bm{\zeta}},{\bm{\chi_j}}\right):\mathcal{F} \times \mathcal{T} \to \{0,1\},
	\end{eqnarray}
	where ${\bm{\chi}}_j=\left[\phi,\psi,{\bm{\varsigma}}\right]^T\in\mathcal{T}$ denotes the parameters of the test function at the $j$th split node. Here, $\psi$ defines the geometric primitive used to separate the input data (e.g., an axis-aligned hyperplane, an oblique hyperplane, a general surface, etc.) \cite{criminisi2012decision}. The parameter vector $
	\bm{\varsigma}$ captures thresholds for
	the inequalities used in the binary test. The filter function $\phi$ selects
	some features of choice out of the entire vector $\bm{\zeta}$. 
	
	At each node $j$, depending on the subset of the incoming training set $\mathcal{S}_j$, we learn the function that ``best'' splits $\mathcal{S}_j$ into $\mathcal{S}_j^L$ and $\mathcal{S}_j^R$. The parameter vector $\bm{\chi}_j$ is selected by maximizing the following objective function at the $j$th node
	\begin{eqnarray}
	\label{eq:goof-rf-eq2}
	\hat{{\bm{\chi}_j}}=\arg\mathop{\max}\limits_{\bm{\chi}_j\in{\mathcal{T}}}\mathcal{I}_j,
	\end{eqnarray}
	where $\mathcal{I}_j$ is called information gain at node $j$ and it is a  function of the vector $\bm{\eta}=\left[\mathcal{S}_j,\mathcal{S}_j^L,\mathcal{S}_j^R,\bm{\chi}_j\right]^T$.
	The left child node $\mathcal{S}_j^L$  and right child node $\mathcal{S}_j^R$ are defined in mathematics as
	\begin{eqnarray}
	\label{eq:goof-rf-eq3}
	\left\{ \begin{array}{l}
	\mathcal{S}_j^L=\left\{\left({\bm{\zeta}},q\right)\in \mathcal{S}_j|h'\left({\bm{\zeta}},{\bm{\chi}}_j\right)=0\right\}
	\\
	\mathcal{S}_j^R=\left\{\left({\bm{\zeta}},q\right)\in \mathcal{S}_j|h'\left({\bm{\zeta}},{\bm{\chi}}_j\right)=1\right\}
	\end{array} \right.,
	\end{eqnarray}
	and the information gain $ \mathcal{I}_j $ can be defined as
	\begin{eqnarray}
	\label{eq:goof-rf-eq4}
	\mathcal{I}_j=\mathcal{H'}\left(S_j\right)-\frac{\left(\left|S_j^L\right|\mathcal{H'}\left(S_j^L\right)+\left|S_j^R\right|\mathcal{H'}\left(S_j^R\right)\right)}{\left|S_j\right|},
	\end{eqnarray}
	where  $\mathcal{H'}\left(S_j\right)$ is the Shannon entropy at node $j$ before the split, which can be defined as
	\begin{eqnarray}
	\label{eq:goof-rf-eq5}
	\mathcal{H'}\left(\mathcal{S}_j\right)=-\sum\nolimits_{{q_i} \in C} {p\left( {{q_i}} \right)} \log _2^{p\left( {{q_i}} \right)},
	\end{eqnarray}
	where $q_i$ indicates the class label of ${\bm{\zeta}_i}$. The set of all classes is denoted as ${C}$ and $ p\left(q_i\right) $ is the empirical  distribution extracted from the  points within the set $\mathcal{S}_j$. Other key model parameters that impact the behavior of a decision tree most include the maximum allowed tree depth $D'$ and the tree number $T'$ in a forest. Given $D'$ , for a binary decision tree, we can calculate the number of internal nodes $ni$ and the number of leaf nodes $nf$ as follows
	\begin{eqnarray}
	\label{eq:goof-rf-eq6}
	\left\{ \begin{array}{l}
	ni={2^{D'} \mathord{\left/
			{\vphantom {1 2}} \right.
			\kern-\nulldelimiterspace} 2}-1\\
	nf={2^{D'} \mathord{\left/
			{\vphantom {1 2}} \right.
			\kern-\nulldelimiterspace} 2}
	\end{array} \right.
	\end{eqnarray}
	The total number of nodes in a decision tree with depth $D'$ is $nr=ni+nf=2^{D'}-1$.

The proposed GOOF-RF algorithm can be divided into an offline training phase and an online testing phase. We first summarize the procedures of our proposed GOOF-RF training algorithm in Algorithm \ref{ALG:multiple_classifiers}.   After obtaining multiple strong classifiers, each testing sample is simultaneously pushed through all trees in these multiple strong classifiers until it reaches the corresponding leaves. Tree testing is done in parallel, thus achieving high computational efficiency. The  GOOF-RF testing algorithm is summarized in Algorithm \ref{ALG:GOOF-RF-TEST}. Note that the function $\textbf{Vote}\left[\cdot\right]$ in Algorithm \ref{ALG:GOOF-RF-TEST} denotes that it chooses the classification having the most votes (over all the trees in the forest). Note that the $q$th entry of the vector $\bm{p}_z$ in Algorithm \ref{ALG:GOOF-RF-TEST} is 1 and others are zeros if the $\gamma$th classifier predicts the location to be $q$. We can transform  $\bm{p}_1,\cdots,\bm{p}_{\mathcal{Z}}$ given by the $\gamma$th strong classifier  into a $\mathcal{Z}\times 1$ vector $\bm{b}_{\gamma}$ whose $z$th entry is the  location label estimated from the $\gamma$th classifier when inputting the $z$th testing sample, i.e.,
	\begin{eqnarray}
	\label{eq:goog-rf-testing}
	\bm{b}_{\gamma}\left(z\right)=\text{loc}\left(\bm{p}_z\neq 0\right),
	\end{eqnarray}
	where $\text{loc}\left(X\right)$ returns the location of nonzero entry in $X$.
	The final output of Algorithm \ref{ALG:GOOF-RF-TEST} is the final prediction matrix $\bm{B}$, as shown in Fig. \ref{fig:MCMSF}.  How to fuse these predictions is the key for indoor localization. In \cite{guo2016localization}, we proposed the MUCUS fusion algorithm to obtain a robust location prediction. However, the timeliness is the bottleneck for real implementation. In this paper, we  derive an improved fusion algorithm to balance the robustness, accuracy, and timeliness.

	\begin{algorithm}[!t]
		\algnewcommand\algorithmicinput{\textbf{Input:}}
		\algnewcommand\algorithmicoutput{\textbf{Output:}}
		\algnewcommand\INPUT{\item[\algorithmicinput]}
		\algnewcommand\OUTPUT{\item[\algorithmicoutput]}
		\caption{\textbf{GOOF-RF Training}}
		\label{ALG:multiple_classifiers}
		\begin{algorithmic}[1]
			\INPUT{
				\begin{inparaenum}
					\item{The training sample set $\mathcal{F} \subset \textbf{GOOF} $.}
					\item{The number of decision trees $T'$ for each fingerprint in the GOOF.}
					\item{The weak learner model $h'\left({\bm{\zeta}},{\bm{\chi}_j}\right)$.}
					\item {The tree depth $D'$.}
					\item {The number of different fingerprints $\mathcal{H}$ in the $\textbf{GOOF}$.}
				\end{inparaenum}
			}
			\OUTPUT {\begin{inparaenum}
					{\textbf{The $\mathcal{H}$ strong random forest classifiers }} ${\bm{H}}$
				\end{inparaenum}}
				\LState {Initiate $\bm{H}\left({{\gamma}}\right)=\emptyset$}
				\For {$\gamma=\{1,\cdots,\mathcal{H}\}$ }
				\LState Select a geometric primitive $\psi$
				\LState Initiate ${\textbf{Tree}}\left(t'\right)=\emptyset$
				\For {$t'=\{1,\cdots,{T'}\}$}
				\LState Compute the number of nodes $nr$ using Eq. (\ref{eq:goof-rf-eq6})
				\LState Set node $\mathcal{S}_0=\emptyset$
				\For {$j=\{1,\cdots,nr\}$}
				\LState Initiate information gain $\mathcal{I}_j$=0
				\LState Select a threshold $\bm{\varsigma}$ randomly based on $\psi$
				\LState Select split dimension $\bm{\phi}\left({\bm{\zeta}}\right)$ based on $\psi$
				\LState Call the weak learner $h'\left({\bm{\zeta}},{\bm{\chi}}_j\right)$
				\LState Compute entropy of node $\mathcal{S}_j^R$ using Eq. (\ref{eq:goof-rf-eq5})
				\LState Compute entropy of node  $\mathcal{S}_j^L$ using Eq. (\ref{eq:goof-rf-eq5})
				\LState Compute information gain $\mathcal{I}_j$  using Eq. (\ref{eq:goof-rf-eq4})
				\LState Choose ${\hat{{\bm{\chi}}}}_j$  at split node $j$ using Eq. (\ref{eq:goof-rf-eq2})
				\LState  ${\textbf{Tree}}\left(t'\right)=\left[{\textbf{Tree}}\left(t'\right) \cup \left(\mathcal{S}_j,{\hat{{\bm{\chi}}}}_j\right)\right]$
				\EndFor
				\LState ${\bm{H}}\left(\gamma\right)=\left[{\bm{H}}\left(\gamma\right) \cup {\textbf{Tree}}\left(t'\right)\right]$
				\EndFor
				\EndFor
				\LState\Return  ${\bm{H}}\left(\gamma\right)$
			\end{algorithmic}
		\end{algorithm}

		\begin{algorithm}[!t]
			\algnewcommand\algorithmicinput{\textbf{Input:}}
			\algnewcommand\algorithmicoutput{\textbf{Output:}}
			\algnewcommand\INPUT{\item[\algorithmicinput]}
			\algnewcommand\OUTPUT{\item[\algorithmicoutput]}
			\caption{\textbf{GOOF-RF Testing}}
			\label{ALG:GOOF-RF-TEST}
			\begin{algorithmic}[1]
				\INPUT{
					\begin{inparaenum}
						\item{The testing sample set ${\bm{\zeta}_z}\in\mathcal{G} \subset \textbf{GOOF} $.}
						\item{The $\mathcal{H}$ strong random forest classifiers $\bm{H}$.}
						\item {The number of testing sample $\mathcal{Z}$.}
					\end{inparaenum}
				}
				\OUTPUT {\begin{inparaenum}
						{\textbf{The prediction matrix  $\bm{B}$}}
					\end{inparaenum}}
					\For {$\gamma=\{1,\cdots,\mathcal{H}\}$ }
					\For {$z=\{1,\cdots,\mathcal{Z}\}$}
					\LState Initiate the prediciton of random forest $\bm{p}={\bm{0}}$
					\For {$t'=\{1,\cdots,{T'}\}$}
					\LState Compute the prediction of the $t'$th tree $\bm{p}_{t'}$
					\EndFor
					\LState $\bm{p}_z=\textbf{Vote}\left[{\bm{p}}_1,\cdots,{\bm{p}}_{T'}\right]$
					\EndFor
					\LState $\bm{b}_{\gamma}=\left[\text{loc}\left(\bm{p}_1\right),\cdots,\text{loc}\left(\bm{p}_{\mathcal{Z}}\right)\right]^T$
					\EndFor
					\LState $\bm{B}=\left[{\bm{b}}_1,\cdots,{\bm{b}}_{\mathcal{H}}\right]$
					\LState\Return  $\bm{B}$
				\end{algorithmic}
			\end{algorithm}
\subsection{Sliding Window aIded Mode-based (SWIM) fusion localization algorithm}
			\label{sec:subsec:mc-mscf}
\begin{figure}[!t]
				\centering
				\includegraphics[width=0.65\columnwidth]{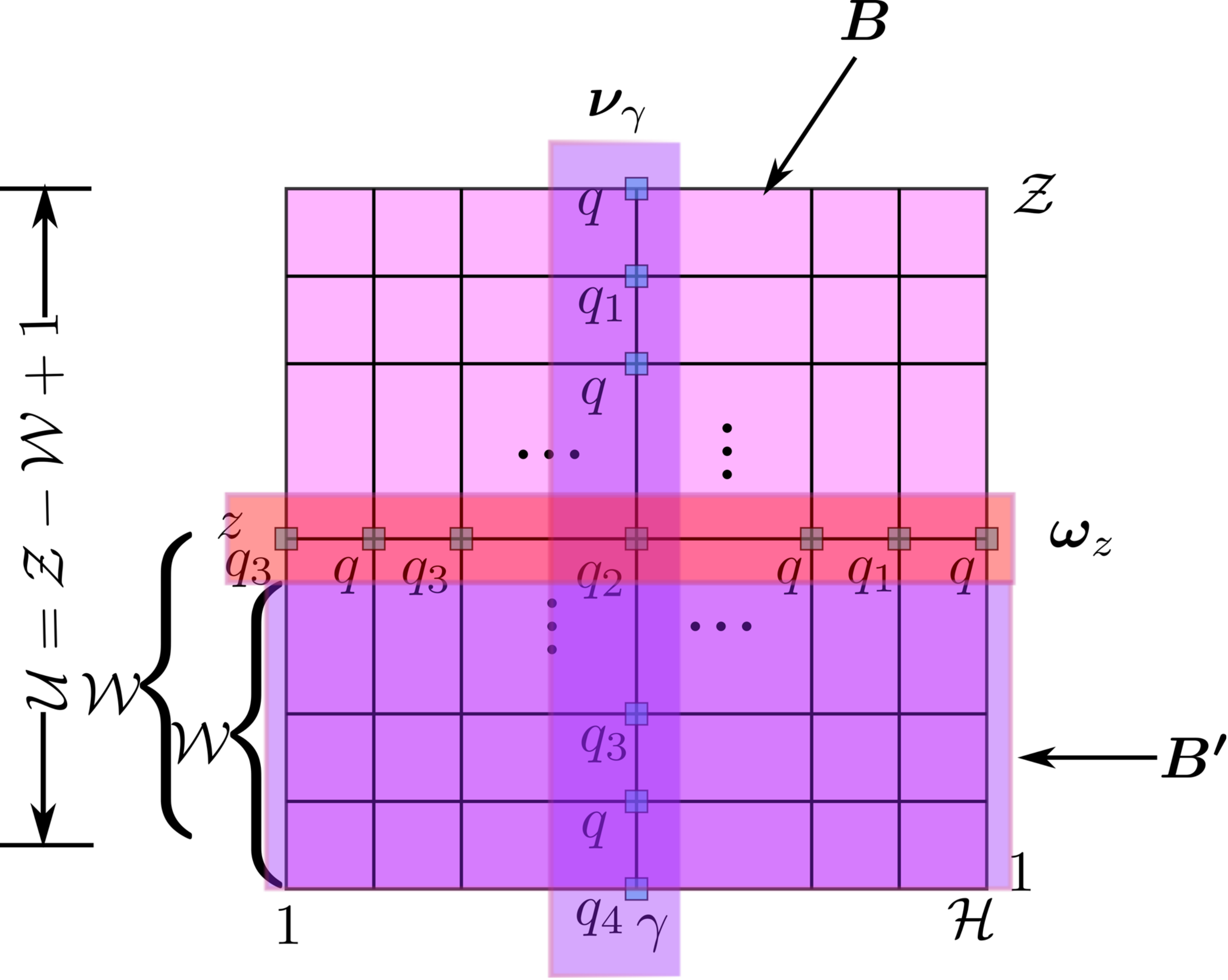}
				\caption{The diagram of the prediction matrix and sliding window.}
				\label{fig:MCMSF}
			\end{figure}

			Let $\mathcal{G}^{\gamma}=\left[\left({\bm{\zeta}}_1^{\gamma},q\right),\left({\bm{\zeta}}_2^{\gamma},q\right),\cdots, \left({\bm{\zeta}}_{\mathcal{Z}}^{\gamma},q\right)\right] \subset \textbf{GOOF} $ be the $\mathcal{Z}$ testing samples at the $q$th grid of the $\gamma$th type fingerprint, where ${\bm{\zeta}}_z^{\gamma}$ is the $z$th sample vector of the $\gamma$th fingerprint and $q$ is the corresponding location label. We can input the $\mathcal{Z}$ testing samples of  the $\gamma$th fingerprint one by one into the $\gamma$th random forest strong classifier, which has been trained by Algorithm \ref{ALG:multiple_classifiers}. Then, the $\gamma$th strong classifier will work as a predictor to give a $\mathcal{Z}\times 1$ prediction vector $\bm{\nu}_{\gamma}$ for the $\mathcal{Z}$ samples, in which  ${\nu}_{\gamma}\left(z\right)$ denotes the output of the $\gamma$th classifier when inputting the $z$th testing sample. The total prediction matrix ${\bm{B}}=\left[\bm{\nu}_1,\cdots,\bm{\nu}_{\mathcal{H}}\right]$, as shown in Fig. \ref{fig:MCMSF}.  For the $z$th testing sample, the $\mathcal{H}$ strong classifiers yield different prediction results $\left\{q, q_1,q_3\right\} \in \bm{\omega}_z$. For the $\gamma$th strong classifier, different testing samples may give different prediction results $\left\{q, q_1,q_2,q_3,q_4\right\} \in \bm{\nu}_{\gamma}$. From these prediction results, we find that the outputs of all the strong classifiers with different samples can be combined to produce a more accurate fusion result. We demonstrate our proposed Sliding Window aIded Mode-based (SWIM) fusion localization algorithm as follows.

First, we can calculate the entropy of the prediction of the $\gamma$th strong classifier with  $\mathcal{Z}$ samples as follows
			\begin{eqnarray}
			\label{eq:mcmscf1-1}
			H\left(\bm{\nu}_{\gamma}\right)=-\sum\limits_{\nu_{\gamma}\left(z\right)\in \mathcal{Q}}p\left(\nu_{\gamma}\left(z\right)\right) \log{p\left(\nu_{\gamma}\left(z\right)\right)}
			\end{eqnarray}
			where $p\left(\nu_{\gamma}\left(z\right)\right)$ is calculated as normalized empirical histogram of predictions of the $\gamma$th strong classifier in $\mathcal{Q}$.
			  $H\left(\bm{\nu}_{\gamma}\right)$ denotes the robustness of the $\gamma$th classifier to the environment noise, i.e., the smaller $H\left(\bm{\nu}_{\gamma}\right)$, the better robustness of the predictions of the $\gamma$th strong classifier.

			Similarly, we can calculate the entropy of the $z$th testing sample for all strong classifiers as follows
			\begin{eqnarray}
			\label{eq:SIAM1}
			H\left(\bm{\omega}_{z}\right)=-\sum\limits_{\omega_{z}\left(\gamma\right)\in \mathcal{Q}}p\left(\omega_{z}\left(\gamma\right)\right) \log{p\left(\omega_{z}\left(\gamma\right)\right)},
			\end{eqnarray}
			where $p\left(\omega_{z}\left(\gamma\right)\right)$ is calculated as the normalized empirical histogram of predictions of the $z$th testing sample in $\mathcal{Q}$.
			Note that $H\left(\bm{\omega}_{z}\right)$ shows the environment adaptability of these fingerprints, i.e., how well these fingerprints cope with multipath and changing environment. The bigger $H\left(\bm{\omega}_{z}\right)$, the more complex of the environment.
			
			Based on the above analysis, a good location estimation should be given from the $\gamma$th strong classifier with the minimal entropy $H\left(\bm{\nu}_{\gamma}\right)$, which has good performance to cope with the environment noise and yields
			\begin{eqnarray}
			\label{eq:SIAM3}
			\hat{\gamma}= \mathop {\min }\limits_{\gamma} H\left(\bm{\nu}_{\gamma}\right),
			\end{eqnarray}
			which means that we choose the predictions of the $\hat{\gamma}$th strong classifier with a high priority which has the best robustness to the  environment noise. In order to combine the predictions of the other classifiers, we first give a mode-based robustness estimator as			
			\begin{eqnarray}
			\label{eq:SIAM31}
			\begin{array}{l}
			\hat q = \text{mode} \left( \bm{B} \right)~~{\rm{subject~to~ }}\hat q \in {{{\bm{\nu}}_{\hat{\gamma}} }},
			\end{array}
			\end{eqnarray}
where $\text{mode}\left(X\right)$  returns the sample mode of $X$, which is the most frequently occurring value in $X$. This estimator means that the optimal estimate must come from the most frequently occurring value in $\bm{B}$, meanwhile, this value must occur in $\bm{\nu}_{\hat{\gamma}}$.
			
			The main drawback of (\ref{eq:SIAM31}) is the timeliness. In general, the value of $\mathcal{Z}$ determines the speed of real localization, the bigger $\mathcal{Z}$, the slower the localization speed. In order to optimize the localization speed and robustness, we further propose the SWIM fusion algorithm as follows.
			
			Assume a rectangular sliding window of length $\mathcal{W}$ ($\mathcal{W} \le \mathcal{Z}$) to be used in the $\mathcal{Z}$ samples. We can just consider fusing $\mathcal{W}\times \mathcal{H}$ submatrice $\bm{B'}$ instead of the total prediction matrix $\bm{B}$. Given a matrix $\bm{B}$, we can obtain some submatrice $\bm{B'}$. The number of $\bm{B'}$ is $\mathcal{U}$. Here, $\mathcal{U}$ is the prediction frequency, i.e., the localization occurrences  per unit time, which shows the speed of localization, $\mathcal{U}=\mathcal{Z}-\mathcal{W}+1$.
			Based on submatrice $\bm{B'}$, we can derive the SWIM algorithm as

			\begin{eqnarray}
			\label{eq:SIAM4}
			\begin{array}{l}
			\hat q = \text{mode} \left( \bm{B'} \right)~~{\rm{subject~to~ }}\hat q \in {{{\bm{\nu}}_{\hat{\gamma}}^{'} }},
			\end{array}
			\end{eqnarray}
			where the submatrix $\bm{B'}$ and vector $\nu_{\hat{\gamma}}^{'}$ are
			\begin{eqnarray}
			\label{eq:SWIM1}
			\bm{B'}=\bm{B}\left(u:u+\mathcal{W},:\right),
			\end{eqnarray}
			and
			\begin{eqnarray}
			\label{eq:SWIM2}
			\bm{{\nu}}_{\hat{\gamma}}^{'}=\bm{{\nu}}_{\hat{\gamma}}\left(u:u+\mathcal{W}\right),
			\end{eqnarray}
			respectively, where $u=1,\cdots, \mathcal{U}$. By using (\ref{eq:SIAM4}), we can obtain a faster localization result with a tiny robustness loss.
			
			Assume that the true grid location is $q$; to evaluate the performance of our proposed algorithm, we define a metric of the prediction  probability $\varrho$ as
			\begin{eqnarray}
			\label{eq:SWIM3}
			\varrho=\frac{\sum\limits_{u=1}^{\mathcal{U}}\left\{\hat{q}=q\right\}}{\mathcal{U}},
			\end{eqnarray}
			which will be used to evaluate the performance of our proposed FAGOT localization framework. The operator $\left\{\hat{q}=q\right\}$ is defined as
			\begin{eqnarray}
			\label{eq:SWIM4}
			\left\{  \hat{q}=q  \right\}{\rm{ = }}\left\{ {\begin{array}{*{20}{c}}
				1\\
				0
				\end{array}{\rm{ }}} \right.\begin{array}{*{20}{c}}
			{{\rm{if }}~~\hat q = q}\\
			{{\rm{if }}~~\hat q \ne q}
			\end{array}
			\end{eqnarray}
\subsection{Performance Analysis}
			\label{sec:subsec:peformance}
			\subsubsection{Localization time}
			
			One of the main advantages of our proposed framework is that the localization time based on $\bm{B'}$ can be shortened to
			${1 \mathord{\left/
					{\vphantom {1 2}} \right.
					\kern-\nulldelimiterspace} \mathcal{U}}$ of the localization time based on $\bm{B}$. The sliding window aided strategy can not only improve the speed of our approach, but also overcome the fluctuation of some SIOF-based localization approaches, which can be seen in experimental results. The larger $\mathcal{W}$, the smaller $\mathcal{U}$ and the slower of the localization speed. However, it can improve the robustness and accuracy of SWIM.
			\subsubsection{Robustness}
			
			Consider the worst case that the predictions of all strong classifiers in a sliding window are different, which corresponds to $\text{mean}\left(H\left(\bm{w}_z\right)\right)=\text{max}\left(H\left(\bm{w}_z\right)\right)$. In this case, if $\bm{\nu}_{\hat{\gamma}}\left(u\right)=\cdots=\bm{\nu}_{\hat{\gamma}}\left(u+\mathcal{W}-1\right)$, which means $\text{min}\left(H\left(\bm{\nu}_{\hat{\gamma}}\right)\right)$ is a small number close to zero, then (\ref{eq:SWIM4}) can give a stable prediction. However, if $\text{min}\left(H\left(\bm{\nu}_{\hat{\gamma}}\right)\right)$ is not the small number close to zero ($H\left(\bm{\nu}_{\hat{\gamma}}\right)$ can be determined in advance depending on  $\mathcal{H}$), it shows that $\bm{\nu}_{\hat{\gamma}}$ has different predictions. So, it may give a wrong prediction based on (\ref{eq:SWIM4}) with a certain probability, and thus leads to the decrease of robustness. In this case, we can improve the robustness by choosing a bigger sliding window length $\mathcal{W}$.
			\subsubsection{Accuracy}
			From (\ref{eq:SWIM4}), we find that our estimator will work well, even if only one of these six strong classifiers works well while the others have  lower prediction probability. Generally speaking, the larger $\mathcal{W}$, the higher accuracy and robustness but slower speed of SWIM. So, how to balance the speed and accuracy is a key problem. In general, $\mathcal{W}$ should be chosen based on the noise level. The basic principle of choosing  $\mathcal{W}$ is that we should choose a larger $\mathcal{W}$ when SNR is lower. In general, if one of the strong classifiers can provide a more robust prediction, $\mathcal{W} \ge 5$ can guarantee that the fusion result of SWIM will not be worse than the best one of these strong classifiers.
			
			\begin{algorithm}[!t]
				\algnewcommand\algorithmicinput{\textbf{Input:}}
				\algnewcommand\algorithmicoutput{\textbf{Output:}}
				\algnewcommand\INPUT{\item[\algorithmicinput]}
				\algnewcommand\OUTPUT{\item[\algorithmicoutput]}
				\caption{\textbf{SWIM}}
				\label{ALG:MC-MSCF}
				\begin{algorithmic}[1]
					\INPUT{
						\begin{inparaenum}
							\item{The {\emph{final prediction matrix}} $\bm{{B}}$.}
							\item{The length of sliding window $\mathcal{W}$.}
						\end{inparaenum}
					}
					\OUTPUT {\begin{inparaenum}
							{\textbf{The prediction probability  $\varrho$ }}.
						\end{inparaenum}}
						\LState Compute $\mathcal{U}=\mathcal{Z}-\mathcal{W}+1$
						\LState Compute $\bm{B'}$ by using (\ref{eq:SWIM1})
						\LState Compute $\bm{\nu'}_{\gamma}$ by using (\ref{eq:SWIM2})
						\For {$u=\{1,\cdots,\mathcal{U}\}$ }
						\For {$\gamma=\{1,\cdots,\mathcal{H}\}$ }
						\LState Compute the entropy  $H\left(\bm{\nu'}_{\gamma}\right)$ by using (\ref{eq:mcmscf1-1})
						\EndFor
						\LState  Find the optimal classifier label $\hat{\gamma}$ by using (\ref{eq:SIAM3})
						\LState  Compute the location estimate $\hat{q}$ by using (\ref{eq:SIAM4})
						\EndFor
						\LState  Calculate the prediction probability by using (\ref{eq:SWIM3})
						\LState\Return  $\varrho$
					\end{algorithmic}
				\end{algorithm}

\section{Experimental Results}
				\label{sec:experimental_results}
We will employ simulation data and real data to test the performance of our proposed algorithms. In the simulation part, we consider different noise by using our proposed signal model in Section \ref{sec:subsec:signal_model}, and in the real experimental setup, we use two SDR \cite{friedman2009angle} platforms to collect real data.
\subsection{Simulation Data}
	Assume we have a ULA of 7 antennas with carrier frequency at 950 MHz. The interspace between adjacent antenna is half wavelength. The uniform PAS model is adopted, i.e., $P_A\left(\theta\right)=
				1 \mathord{\left/
					{\vphantom {1 {\left( {2\sqrt 3 {\sigma _A}} \right)}}} \right.
					\kern-\nulldelimiterspace} {\left( {2\sqrt 3 {\sigma _A}} \right)}$, where $\sigma _A$  is defined by (\ref{eq:signal_model5}). An $8 \rm{m} \times  8 \rm{m}$ indoor environment is divided into $\mathcal{Q}=64$ grids with equal interspace of 1m. The location of the $q$th grid is denoted as $[x_q,y_q,z_q]$; the ULA is deployed at the corner of this room with the location of the central element being $\left[0,0,z\right]$,  and its normal direction points to the diagonal of the indoor area (for simplicity, we just consider a 2D indoor environment, i.e., $z$=0). The CAoA  $\theta_0$  and average time delay $\tau_0$ of the transmitted signal are calculated from the locations of receiver array and the $q$ location $[x_q,y_q]$. We control the  time delay spread (DS) and angular spread (AS) to be $\tau_0/10$ and $25^{\circ}$, respectively. We add 20 paths to each LOS at each grid.
				
First,  Gaussian white noise is added to the generated signals.  The signal-noise-ratio (SNR) is defined as ${\rm{SNR}}=10\log10^{\frac{\sigma_s^2}{\sigma_n^2}}$, where $\sigma_s^2$ and $\sigma_n^2$ are signal and noise variance, respectively. The total number of snapshots is 3200  at each grid, and we get $\mathcal{Z}=100$ samples with each sample having 32 snapshots. The SNRs are set from -10 to 30 dB with 8 dB interspace. The $\mathcal{H}=6$  fingerprints are considered. We build the GOOF by using Algorithm \ref{ALG:FsGbuilding}, and then we divide each of these  fingerprints into two groups: one with 60 samples is used to train the multiple strong classifiers, and the other with 40 samples is used to test these classifiers. The tree number of each random forest $T'=50$, the tree depth $D'= 8$, and the weak learner $h'$ is a decision dump.		
				
\begin{figure}[!t]
	\centering
	\includegraphics[width=0.65\columnwidth]{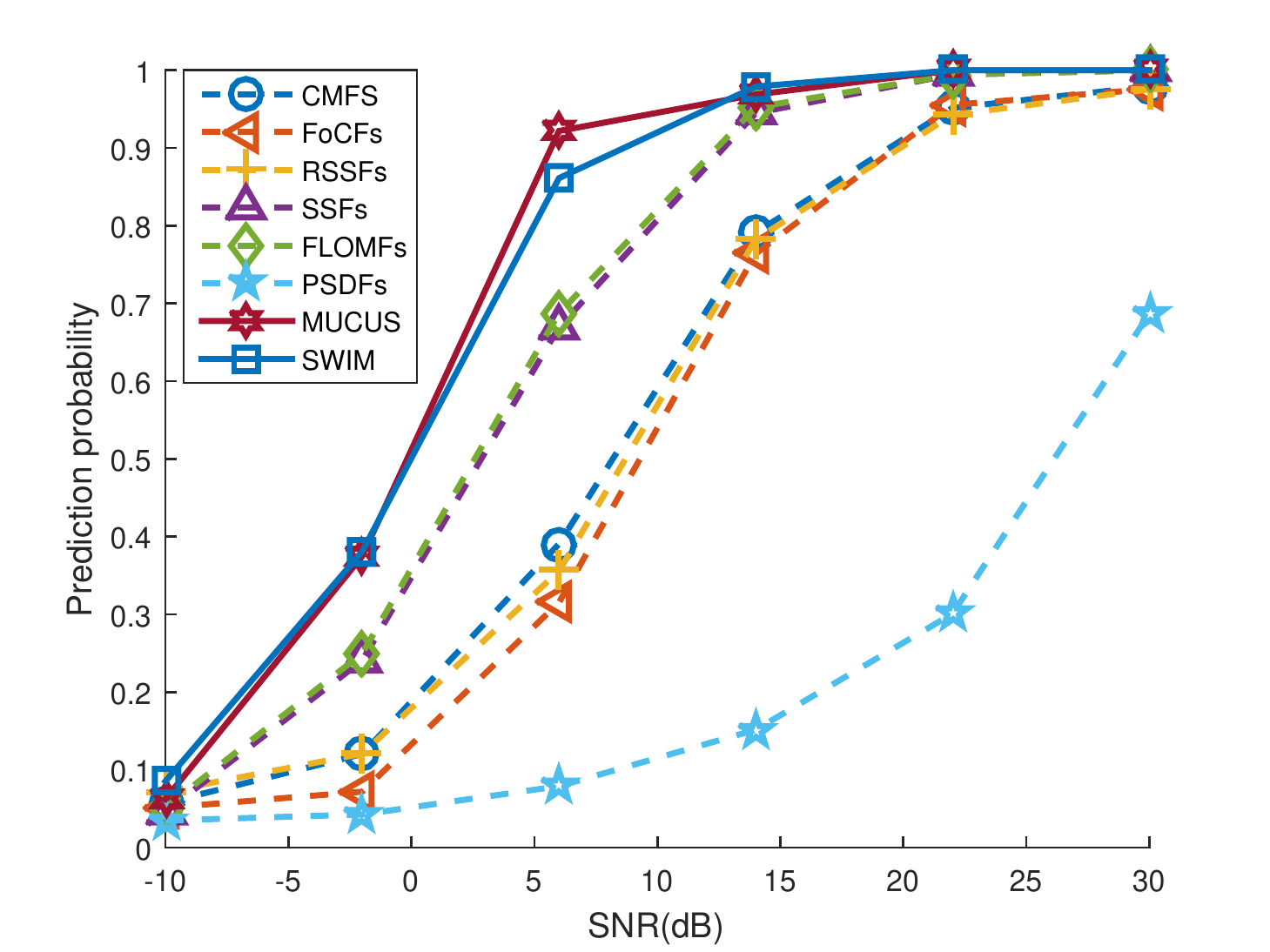}
	\caption{The prediction performance of different algorithms versus different SNRs: Gaussian noise.}
	\label{fig:mc-mscf}
\end{figure}

Fig. \ref{fig:mc-mscf} shows the average prediction probabilities for all $64$ grids of  SWIM  versus different SNRs. In this figure, the CMFs, PSDFs, FoCFs, RSSFs, FLOMFs, and SSFs are the prediction results by using random forests separately. The MUCUS curve is calculated from all 40 testing samples at all grids \cite{guo2016localization}. The curve of our proposed SWIM algorithm is obtained by using a sliding window with length $\mathcal{W}=5$. It is seen that SWIM and MUCUS have higher prediction probabilities than the other SIOF-based methods regardless of the SNRs. SWIM almost has the same performance as MUCUS. However, the prediction frequency of SWIM is $\mathcal{U}=40-\mathcal{W}+1=36$, while the prediction frequency of MUCUS is $1$. This means that SWIM can give 36 times location predictions, but MUCUS can just produce one prediction result in the same time period. Hence, SWIM is a faster algorithm as compared with MUCUS.

\begin{figure}[!t]
\centering
\includegraphics[width=0.65\columnwidth]{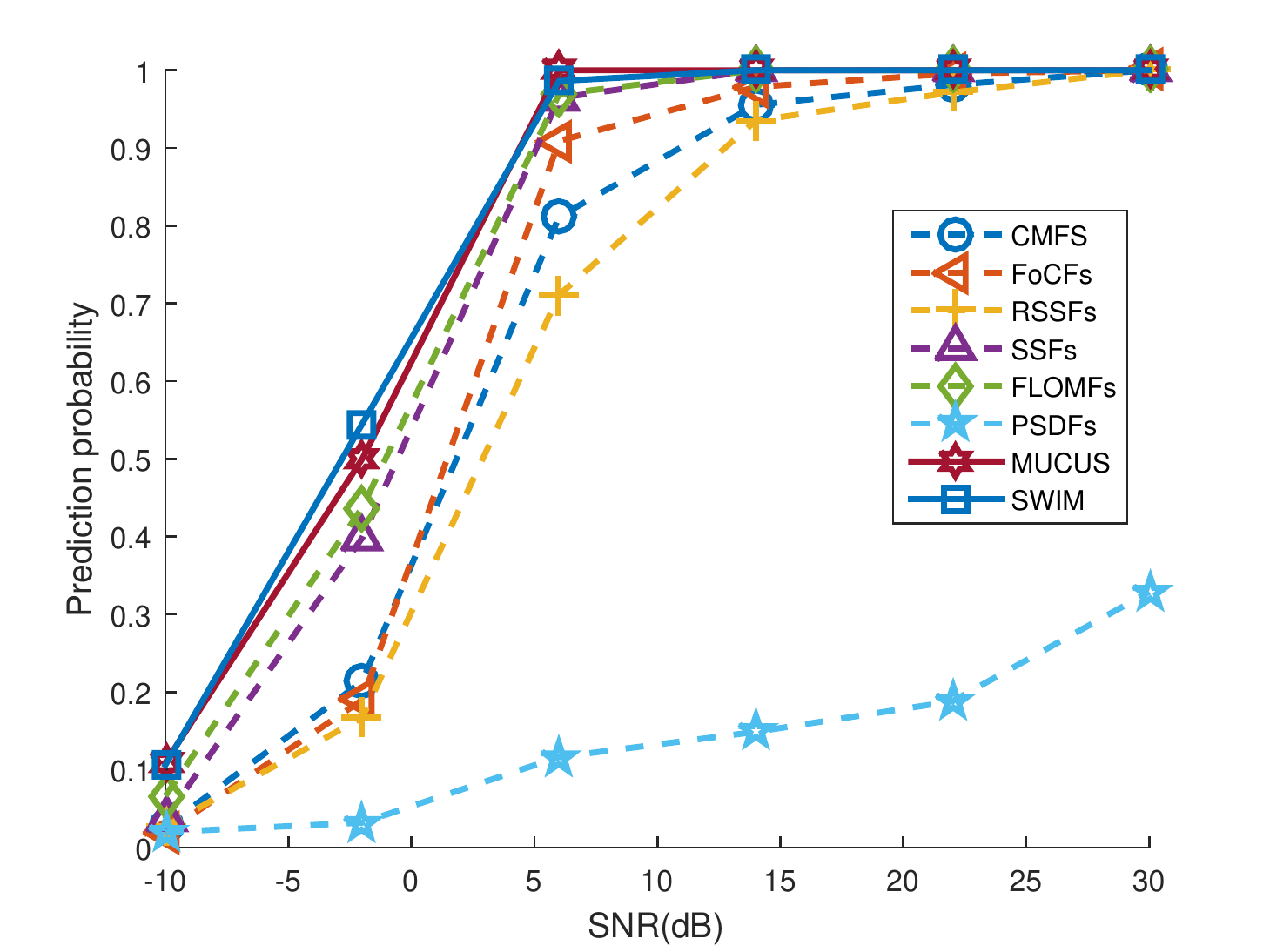}
\caption{The prediction performance of different algorithms versus different SNRs: color noise.}
\label{fig:mc-mscf-color}
\end{figure}

Second, we consider the color noise case. The color noise is generated from filtering Gaussian white noise by using a finite impulse response (FIR) filter with a rectangular window of length 5. The same SNRs are considered. The parameters of random forest and sliding window are the same as the ones in above the Gaussian noise case. The results of these algorithms are illustrated in Fig. \ref{fig:mc-mscf-color}. As shown in Fig. \ref{fig:mc-mscf-color}, SWIM shows  better performance than MUCUS when SNRs are low. Both of them are better than the other six SIOF-based localization algorithms.
				
\begin{figure}[!t]
\centering
\includegraphics[width=0.65\columnwidth]{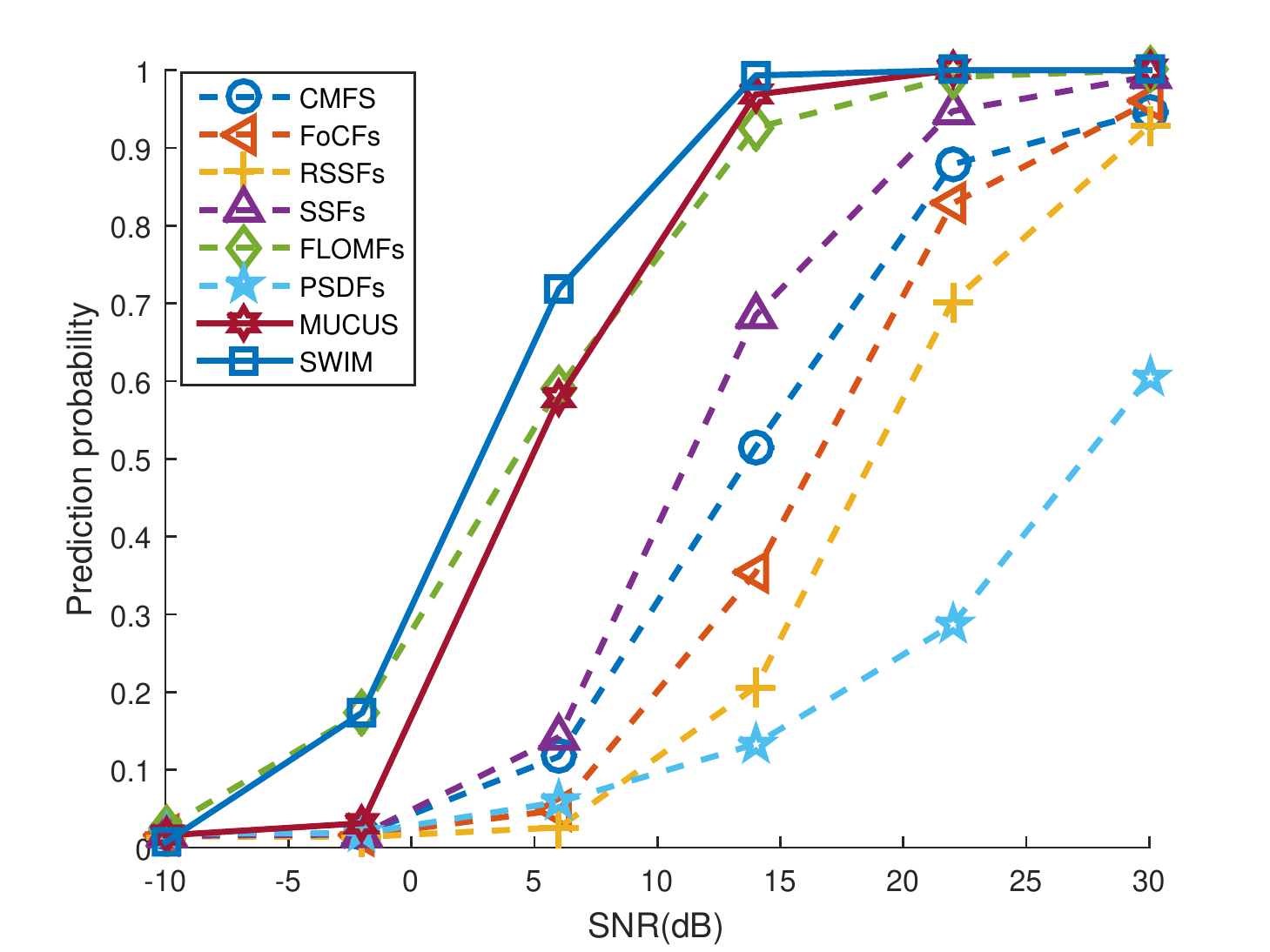}
\caption{The prediction performance of different algorithms versus different SNRs: impulse noise.}
\label{fig:mc-mscf-impulse}
\end{figure}

\begin{figure}[!t]
\centering
\includegraphics[width=0.65\columnwidth]{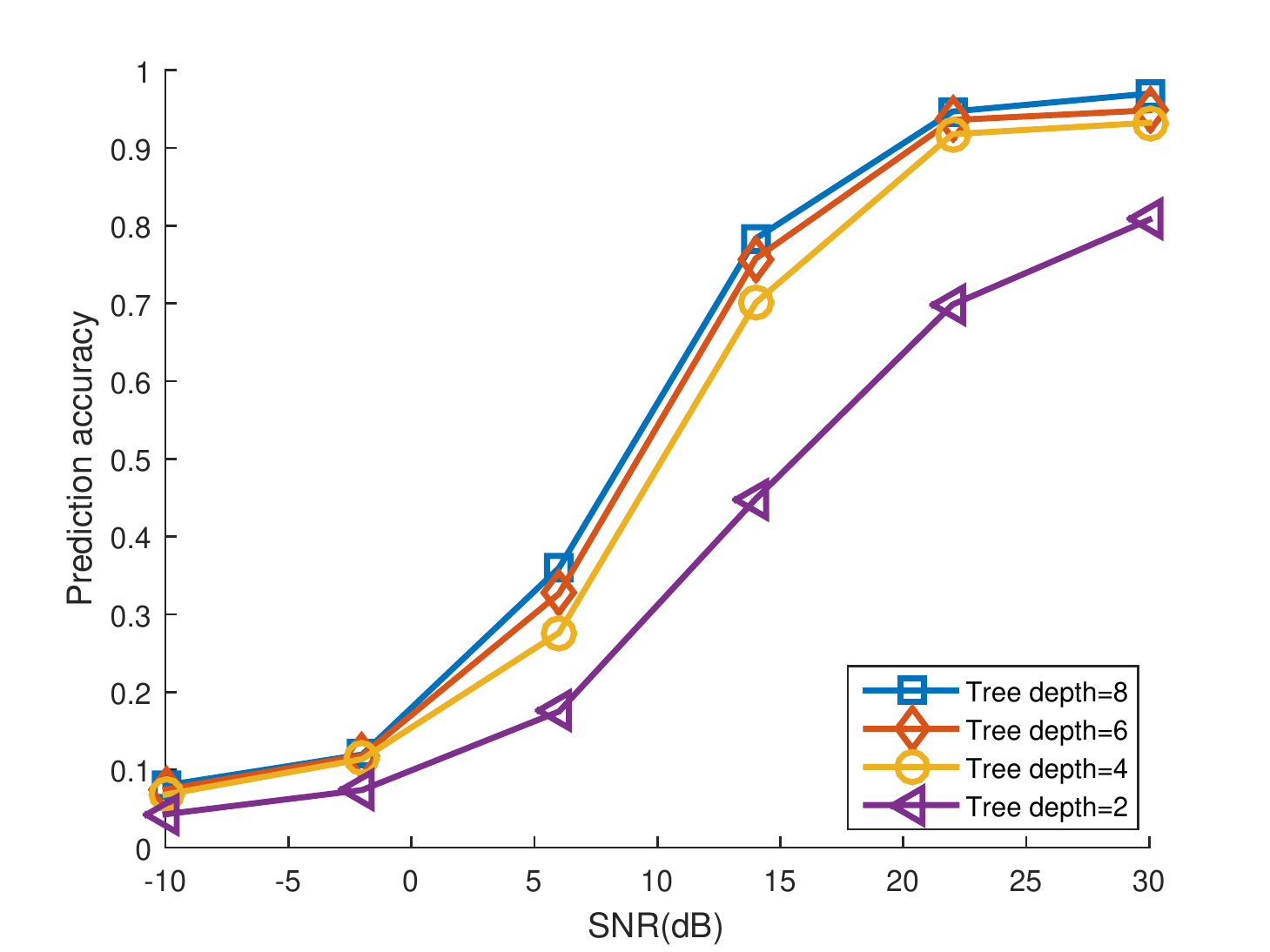}
\caption{The prediciton probability versus SNRs with different tree depths.}
\label{fig:mc-mscf-treedepth}
\end{figure}

\begin{figure}[!t]
\centering
\includegraphics[width=0.65\columnwidth]{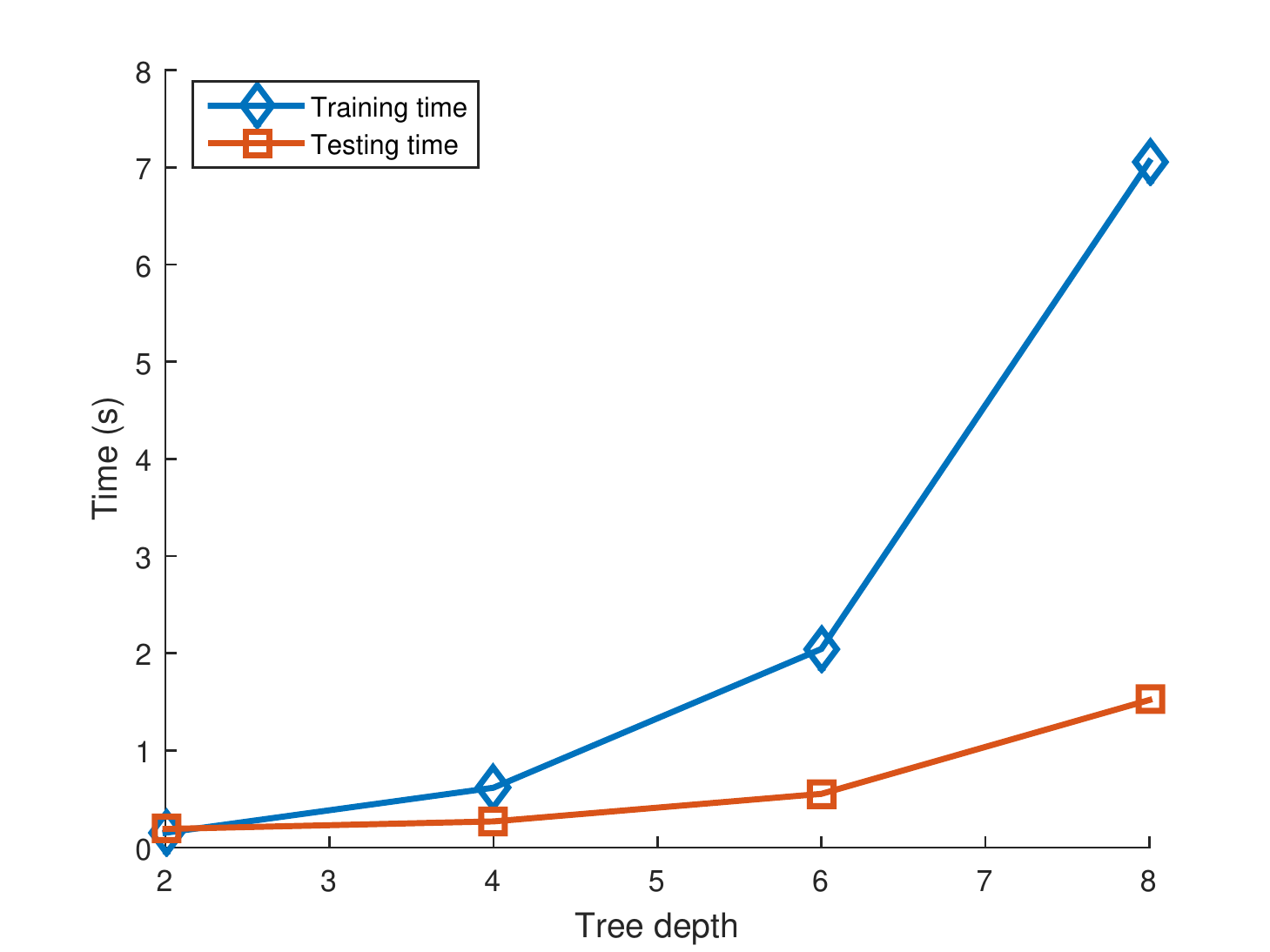}
\caption{The training and testing time versus different tree depths.}
\label{fig:mc-mscf-time-depth}
\end{figure}

Now, we consider an impulse noise case. A (S$\alpha$S) processes whose SNR is defined as ${\rm{SNR}}=10\log\left(\frac{E\left\{s^2\left(t\right)\right\}}{\xi}\right)$, where $\xi$ is the dispersion parameter. The other parameters are $\alpha=1.4$, $\beta=0$, $\delta=0$ \cite{zhong2013particle}. The same SNRs values are considered. Fig. \ref{fig:mc-mscf-impulse} shows the prediction results of SWIM versus different SNRs. As compared with MUCUS and the other six SIOF-based algorithms, SWIM obtains the best predictions for different levels of impulse noise. We can conclude from Figs. \ref{fig:mc-mscf},  \ref{fig:mc-mscf-color}, and \ref{fig:mc-mscf-impulse} that SWIM and MUCUS have nearly the same performance in cases with higher SNRs, while SWIM shows higher accuracy in cases with lower SNRs, such as SNRs below zero because the exponent weighting strategy used in MUCUS cannot select a correct prediction when all the strong classifiers show poor performance. SWIM cannot work well in an extremely poor case too, such as SNR=-10 dB as shown in Fig. \ref{fig:mc-mscf-impulse} because all classifiers cannot work well in this case.
To test the performance of GOOF-RF versus different random forests parameters. At each grid, we choose 50 RSS samples randomly in the GOOF as the training fingerprints, and the other 50 samples as the testing fingerprints. We evaluate the prediction probability versus different $D'$ in Fig. \ref{fig:mc-mscf-treedepth}. Here, $T'=50$ and $h'$ is a decision stump. Note that the prediction probability becomes better as $D'$ increases from 2 to 8. However, the performance shows limited improvement when $D'$ increases from 6 to 8, while the training time and testing time  become much longer, as shown in Fig. \ref{fig:mc-mscf-time-depth}.
				
\begin{figure}[!t]
\centering					
\includegraphics[width=0.65\columnwidth]{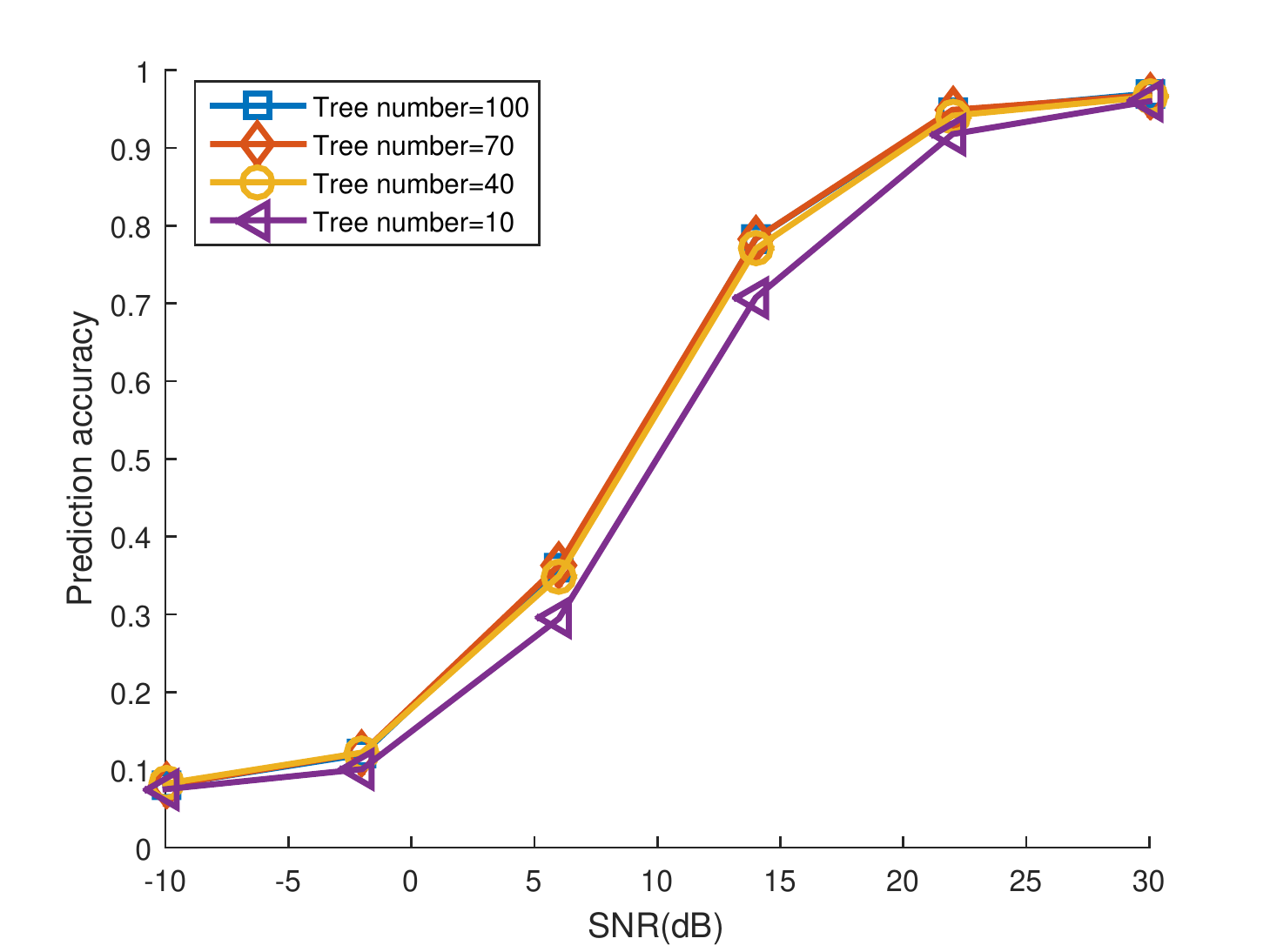}
\caption{The prediciton probability versus SNRs with different tree numbers.}
\label{fig:mc-mscf-treenumber}
\end{figure}

\begin{figure}[!t]
\centering
\includegraphics[width=0.65\columnwidth]{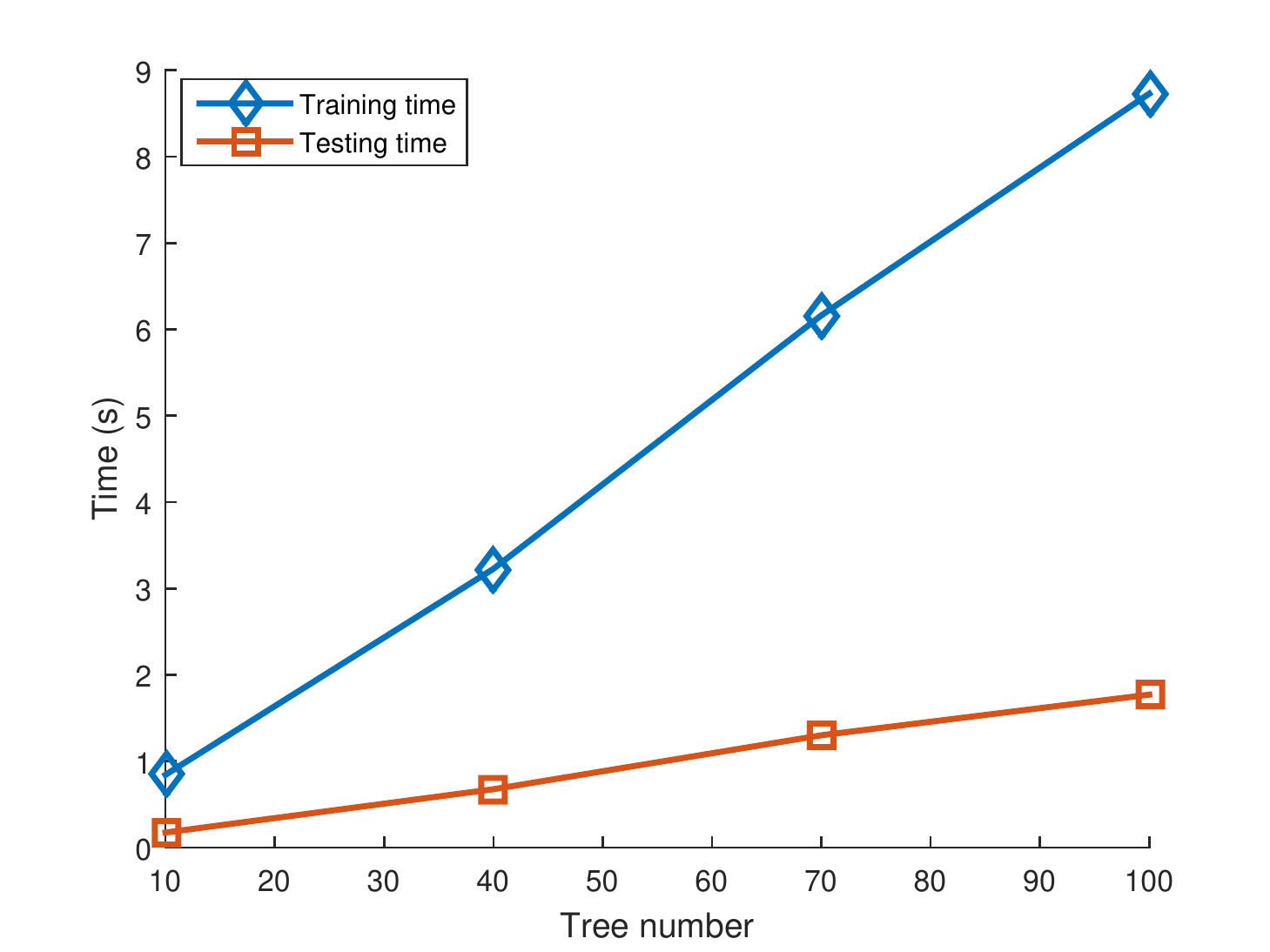}
\caption{The training and testing time versus different tree numbers.}
\label{fig:mc-mscf-time-number}
\end{figure}

			 We fix  $D'=8$ and $h'$ to be the above decision dump and change  $T'$ from 10 to 100 with 30 interspaces. We depict the average prediction probability versus SNRs with different $T'$ in Fig. \ref{fig:mc-mscf-treenumber}. The corresponding average training and testing time are shown in Fig. \ref{fig:mc-mscf-time-number}. It is seen that the best $T'$ for our problem is 40, too large $T'$ cannot improve the prediction probability significantly. Note that the training time belongs to the offline phase, which does not effect the localization speed. We here just evaluate the testing time. From Fig. \ref{fig:mc-mscf-time-number}, we can find the total time of testing 40 samples is $0.6751$s when $T'=40$ and $D'= 8$. Hence, the testing time for each sample is about ${{0.6751} \mathord{\left/
						{\vphantom {{0.6751} {40}}} \right.
						\kern-\nulldelimiterspace} {40}} = 0.0169$s. This can be treated as the localization time of SWIM because the sliding window we used can obtain each prediction corresponding to every input testing sample after the first 5 samples. While MUCUS needs $0.6751$s to give a prediction. So, SWIM algorithm is much faster than MUCUS.

\subsection{Real Data}
We use the SDR technology to build our testbeds. The experiment receiver platform is based on two Universal Software Radio Peripheral1 (USRP1) units; each USRP1 is equipped two RFX900 daughterboards and each daughterboard is equipped two antennas (i.e, a total of four antenna elements), and  the transmitter platform is one USRP1 with one RFX900 daughterboard and one antenna. The transmitter and the receiver platforms are developed based on the open-source software toolkit named GNU Radio. The operating system is Ubuntu10.10. The pictures of this two experimental platforms can be seen in  \cite{guo2016localization}.
	The experimental environment is the KB508 laboratory at University of Electronic Science and Technology of China (UESTC),  which has many desks, partitions, and about 30 graduate students. The topological layout is also shown in \cite{guo2016localization}. The length and width of our laboratory are 9.8m and 6.3m, respectively. The receiver array with 4 antennas is deployed at the corner of the laboratory at the height of 1.5m.
				
\begin{figure}[!t]
\centering
\includegraphics[width=0.65\columnwidth]{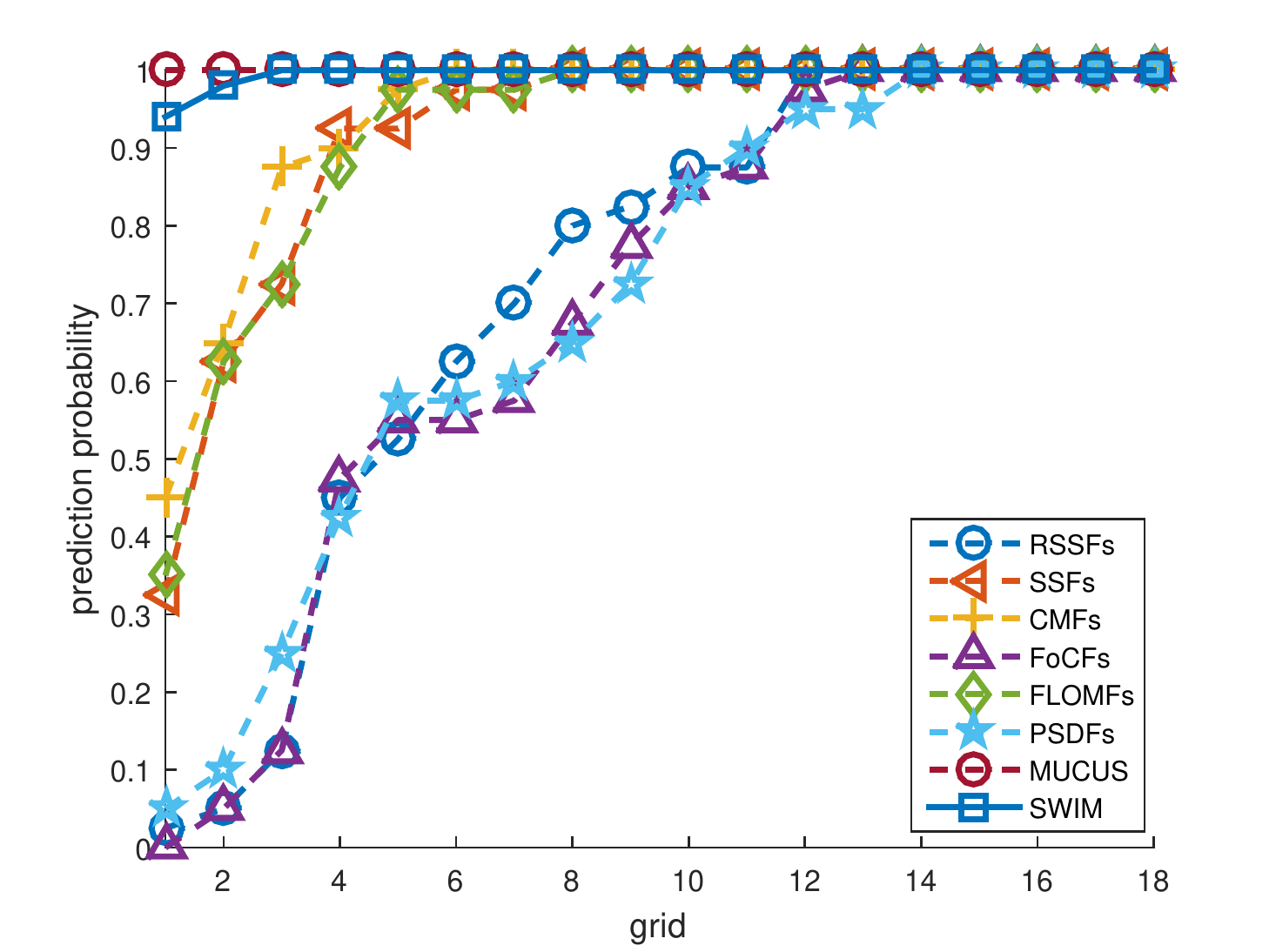}
\caption{The comparison of prediction probability at each grid.}
\label{fig:mc-mscf_grid}
\end{figure}

We transmit a cosine signal with carrier frequency of 900MHz at 18 grid to build the GOOF by using the signals received at the four antennas. $L=400$ snapshots are taken, and are divided into $\mathcal{M}=80$ samples with each group having 5 snapshots, i.e., we just use 5 (${L \mathord{\left/{\vphantom {L M}} \right.\kern-\nulldelimiterspace} \mathcal{M}}=5$) snapshots to estimate each fingerprint at each grid.  Each kind of fingerprints has 80  samples incorporated in our final GOOF. We use 40 samples (${\mathcal{M} \mathord{\left/{\vphantom {L M}} \right.
\kern-\nulldelimiterspace} 2}=40$) as the training data, and the other 40 ($\mathcal{Z}=40$) samples as the testing data. $h'$ is a 2D general oriented hyperplane; $D'= 8$, $T'=50$. A sliding window of length $\mathcal{W}=10$ is adopted to the 40 testing samples. So, we can obtain $40-\mathcal{W}+1=31$ predictions at each grid.   The prediction probability of SWIM is calculated based on the $31$ predictions by using (\ref{eq:SWIM3}), as shown in Fig. \ref{fig:mc-mscf_grid}. While MUCUS uses all the 40 samples to give one prediction. From Fig. \ref{fig:mc-mscf_grid}, it shows that the prediction probabilities of MUCUS at all grids are one because it gives only one prediction at each grid and all the predictions are correct. SWIM can obtain almost the same performance as MUCUS even with the shorter sliding window Note that our results are given without any knowledge of environment. So, our algorithm is very robust to the unknown indoor environment. By the way, the number of snapshots used to compute the GOOF is only five, which may degrade the fingerprints in the GOOF. However, our proposed fusion algorithm can work well in this case.
				
\begin{figure}[!t]
\centering
\includegraphics[width=0.65\columnwidth]{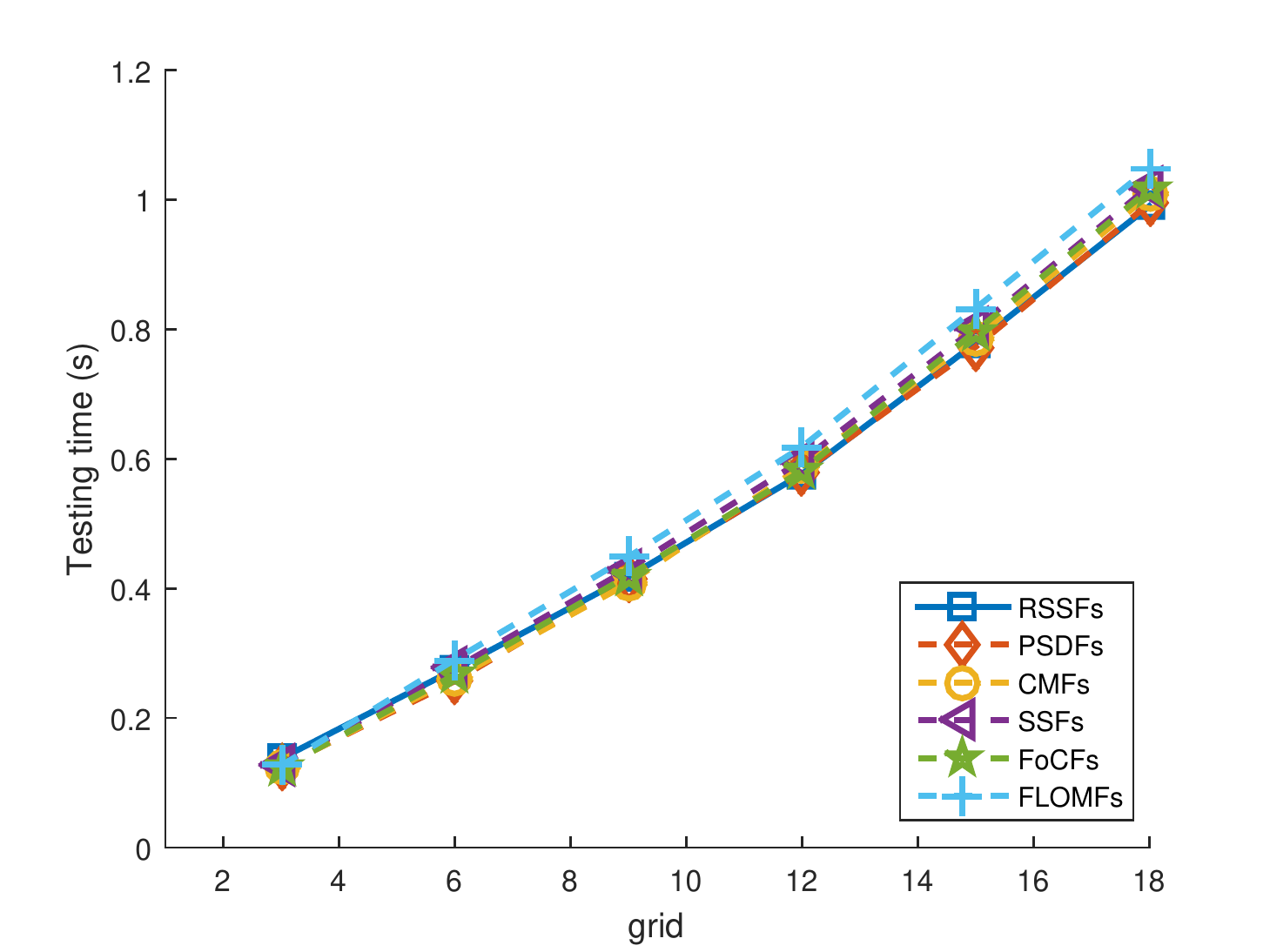}
\caption{The testing time versus different grid numbers.}
\label{fig:testing_time}
\end{figure}

Fig. \ref{fig:testing_time} shows the evaluation of the localization speed of our proposed algorithm, with the testing time of 40 samples versus different grid numbers. The testing time of different fingerprints shows little differences. From this figure, we find that the total testing time of 18 grids is about 1.0231s. So, the testing time of each grid is about 0.0568s. In this time, MUCUS can only output one location prediction, while SWIM can produce 31 location predictions with $\mathcal{W}=10$. So, our proposed algorithm can balance the localization speed and robustness well. 				

\section{Conclusion}
\label{sec:con}
Based on the FAGOT localization framework, we have proposed random forest based classifiers training and testing algorithms. We have also proposed  SWIM  to balance the speed and robustness of indoor localization. Our proposed algorithms can not only overcome drawbacks of SIOF-based indoor localization methods, but also balance the robustness and speed of the MUCUS \cite{guo2016localization}.
				
Apart from the fingerprints constructed in the GOOF of \cite{guo2016localization}, we have also incorporated PSDFs into our GOOF, which further enrich the channel information of the GOOF. Of course, other fingerprints, such as CIR, PDDP, crowdsourcing, can also be added into the GOOF. All in all, GOOF strategy just represents the developing trend of building fingerprints. Here, we just discuss random forest based classifiers and AdaBoost based classifiers \cite{guo2016localization}. Other machine learning methods, such as Support Vector Machine (SVM), Neural Network (NN), and Convolutional Neural Network (CNN), can also be used as classifiers. How to  use this GOOF better by using  machine learning is an interesting future pursuit.
				
Simulations show that our proposed localization framework can achieve better performance regardless of noise types. The real experiment results demonstrate that our proposed localization framework is still robust to the unknown localization environment. The proposed GOOF can not only reduce the burden of building fingerprints,  but also can offer more  information about the indoor environment for further localization, and is thus very attractive for localization in an unknown complex indoor environments.

\ifCLASSOPTIONcaptionsoff
  \newpage
\fi

\bibliographystyle{IEEEtran}
\bibliography{gxs_fingerprint}
\end{document}